\documentclass[lettersize,journal]{IEEEtran}
\usepackage{amsmath,amsfonts}
\usepackage{algorithmic}
\usepackage{algorithm}
\usepackage{array}
\usepackage[caption=false,font=normalsize,labelfont=sf,textfont=sf]{subfig}
\usepackage{textcomp}
\usepackage{stfloats}
\usepackage{url}
\usepackage{verbatim}
\usepackage{graphicx}
\usepackage{cite}
\usepackage{xcolor} 
\usepackage{graphicx} 
\usepackage{booktabs}
\usepackage{threeparttable}

\begin{document}

\title{Dual Modality Prompted Diffusion Priors \\
for Zero Shot Hyperspectral Pansharpening}




\author{Pengwei~Xie, 
        Fei~Zhu, 
        Jiajun~Li, 
        Xiangyuan~Liu, 
        Kangqing~Shen, 
        and~Gemine~Vivone,~\IEEEmembership{Senior~Member,~IEEE}% <-this % stops a space
% \thanks{Manuscript received XX XX, 2026; revised XX XX, 2026. P. Xie and F. Zhu contributed equally to this work. \textit{(Corresponding author: Kangqing Shen.)}}% 

\thanks{P. Xie and F. Zhu contributed equally to this work. \textit{(Corresponding author: Kangqing Shen.)}}% <-this % stops a space

\thanks{P. Xie is with the School of Artificial Intelligence, Beijing Normal University, Beijing 100875, China (e-mail: pwxie@mail.bnu.edu.cn). 
F. Zhu is with the School of GeoAI and Hinton STAI Institute, and the Key Laboratory of Geographic Information Science (Ministry of Education), East China Normal University, Shanghai 200241, China (e-mail: 52290161013@stu.ecnu.edu.cn). 
J. Li is with Peking University, Beijing 100871, China (e-mail: jiajun.lee@stu.pku.edu.cn). 
X. Liu is with the National Center for Applied Mathematics Shenzhen (NCAMS), Southern University of Science and Technology, Shenzhen 518055, China (e-mail: liuxy7@mail.sustech.edu.cn). 
K. Shen is with the Department of Automation, Tsinghua University, Beijing 100084, China (e-mail: shenkq@mail.tsinghua.edu.cn). 
G. Vivone is with the National Research Council of Italy, Institute of Integrated Methodologies for Earth Observation (CNR-IMIOT), Tito 85050, Italy (e-mail: gemine.vivone@imaa.cnr.it).}}

% \IEEEpubid{0000--0000/00\$00.00~\copyright~2026 IEEE}

% \IEEEpubid{0000--0000/00\$00.00~\copyright~2026 IEEE}

% \markboth{IEEE Transactions on Geoscience and Remote Sensing,~Vol.~XX, No.~XX, XX~2026}%
\markboth{}%
{Xie \MakeLowercase{\textit{et al.}}: Dual Modality Prompted Diffusion Priors for Zero Shot Hyperspectral Pansharpening}

% \IEEEpubid{0000--0000/00\$00.00~\copyright~2026 IEEE}

\maketitle

\begin{abstract}
Hyperspectral pansharpening aims to reconstruct a high resolution hyperspectral (HRHS) image from a panchromatic (PAN) image and a low resolution hyperspectral (LRHS) image while preserving both spatial details and spectral fidelity. Recent diffusion based methods exploit pretrained image priors by generating a low dimensional representation and subsequently mapping it to the hyperspectral domain. However, the observed panchromatic and hyperspectral images are typically imposed only through external reconstruction objectives, limiting their direct interaction with the diffusion prior. To address this issue, we propose dual-modality image-prompted diffusion model (DIDM) for zero shot hyperspectral pansharpening. DIDM encodes the low resolution hyperspectral and panchromatic observations into spectral and spatial prompt tokens, respectively, and injects them into intermediate features of a frozen remote sensing diffusion model through cross attention, allowing complementary spectral and spatial information to directly guide diffusion feature evolution. In addition, we introduce a panchromatic guided weighted pixel aware total variation regularizer that combines low resolution hyperspectral degradation fidelity and panchromatic response fidelity with gradient adaptive structural regularization, thereby preserving structural discontinuities while suppressing spurious variations in homogeneous regions. Extensive experiments on Pavia, Chikusei, and Houston under reduced resolution protocols show that DIDM achieves the best performance across all evaluated metrics, while full resolution evaluation on FR1 yields the highest HQNR among the compared methods. These results demonstrate that internal dual modality prompting and panchromatic guided structural regularization provide an effective balance between spatial detail enhancement and spectral preservation.
\end{abstract}


\section{Introduction}

\IEEEPARstart{H}{yperspectral} pansharpening refers to fusing a low-resolution hyperspectral (LRHS) observation and a high-resolution panchromatic (PAN) image to reconstruct a high-resolution hyperspectral (HRHS) image. This task is an important preprocessing step for downstream hyperspectral image interpretation, because the reconstructed HRHS is expected to contain both high-resolution spatial details and discriminative spectral information \cite{Loncan2015Review,Vivone2015Critical}. Such data support intelligent perception tasks, including thematic classification and material discrimination, in which joint spatial--spectral modeling is essential \cite{Li2024MambaHSI,Yang2026FedDAHSI}. They are also valuable for urban and environmental analysis, such as land-cover mapping, where clear spatial boundaries and faithful spectral responses are critical \cite{Roy2024CrossHL,Lin2025DCMNet}. Unlike multispectral pansharpening, hyperspectral pansharpening must recover many contiguous spectral bands from observations with mismatched spatial and spectral resolutions. Its central challenge is therefore to enhance PAN-guided spatial details without distorting the spectral characteristics inherited from LRHS.

Existing methods can be divided into model-based and learning-based approaches. Model-based methods rely on explicit observation models and hand-crafted priors: GSA injects PAN details through adaptive component substitution \cite{Aiazzi2007GSA}, CNMF reconstructs HRHS through coupled matrix factorization \cite{Yokoya2012CNMF}, HySure formulates hyperspectral super-resolution as a convex subspace-regularized inverse problem \cite{Simoes2015HySure}, and TV-based methods impose gradient regularization to preserve structures and suppress artifacts \cite{Addesso2017CTV,Takeyama2020HSSTV}. Although these methods are interpretable and training-free, their fixed priors have limited flexibility in modeling complex textures and nonlinear spatial--spectral interactions across diverse scenes. Learning-based methods improve representation capacity by learning nonlinear mappings from LRHS/PAN observations to HRHS. Supervised Convolutional Neural Network (CNN)-based, attention-based, and deep-unfolding networks achieve competitive performance under simulated training protocols \cite{Dian2018DHSIS,Guo2022EIA,Xie2022MHFNet}, while deep-prior and spatial--spectral attention networks improve detail injection and spectral preservation \cite{Han2020DHPDARN,Hu2022SSANet,Liu2022SSAU}. Recent hyperspectral fusion networks employ selective re-learning to focus computation on spatially or spectrally degraded features \cite{Liu2025SRLFNet}. However, paired HRHS labels are difficult to acquire in real satellite imaging, and training on synthetically degraded data remains dependent on the assumed degradation model \cite{Ciotola2025CriticalReview}. Recent zero-shot methods reduce this dependence by adapting directly to the target observations. ZSL performs hyperspectral super-resolution without external paired supervision \cite{Qu2022ZSL}; more recently, $\rho$-PNN introduces band-wise adaptation with hysteresis-based spectral-quality control \cite{Guarino2025RhoPNN}, while Hipandas jointly addresses pandenoising and pansharpening through a zero-shot restoration framework with guided reconstruction networks and low-rank priors \cite{Xu2025Hipandas}. Nevertheless, without a sufficiently expressive image prior, per-image optimization can struggle to recover realistic high-frequency spatial structures.

Diffusion priors provide a possible way to strengthen per-image reconstruction. Diffusion models learn iterative denoising processes and serve as powerful image priors \cite{Ho2020DDPM,Song2021DDIM,Nichol2021ImprovedDDPM,Dhariwal2021DiffusionBeatGANs}. They have also shown strong potential in restoration and inverse problems, including super-resolution, inpainting, latent high-resolution synthesis, and posterior sampling \cite{Saharia2023SR3,Lugmayr2022RePaint,Rombach2022LDM,Kawar2022DDRM,Chung2023DPS}. Directly applying an available pre-trained diffusion model to full-band HRHS is nevertheless nontrivial, because most pre-trained priors operate in RGB or another low-dimensional image domain rather than in a high-dimensional hyperspectral space. Representative hyperspectral diffusion methods therefore perform reverse sampling in a low-dimensional image space and subsequently recover the full spectral image. HIR-Diff combines a pre-trained diffusion prior with a low-rank representation for unsupervised hyperspectral restoration \cite{Pang2024HIRDiff}. PLRDiff adapts low-rank diffusion modeling to hyperspectral pansharpening \cite{Rui2024PLRDiff}. DM-ZS advances zero-shot pansharpening by coupling iterative zero-shot guidance with neural spatial-spectral decomposition (NSSD) reconstruction for a given LRHS/PAN pair \cite{Xiao2025DMZS}. These studies demonstrate the value of pre-trained diffusion priors when paired HRHS supervision is unavailable.

\begin{figure*}[t]
    \centering
    \includegraphics[width=0.98\textwidth]{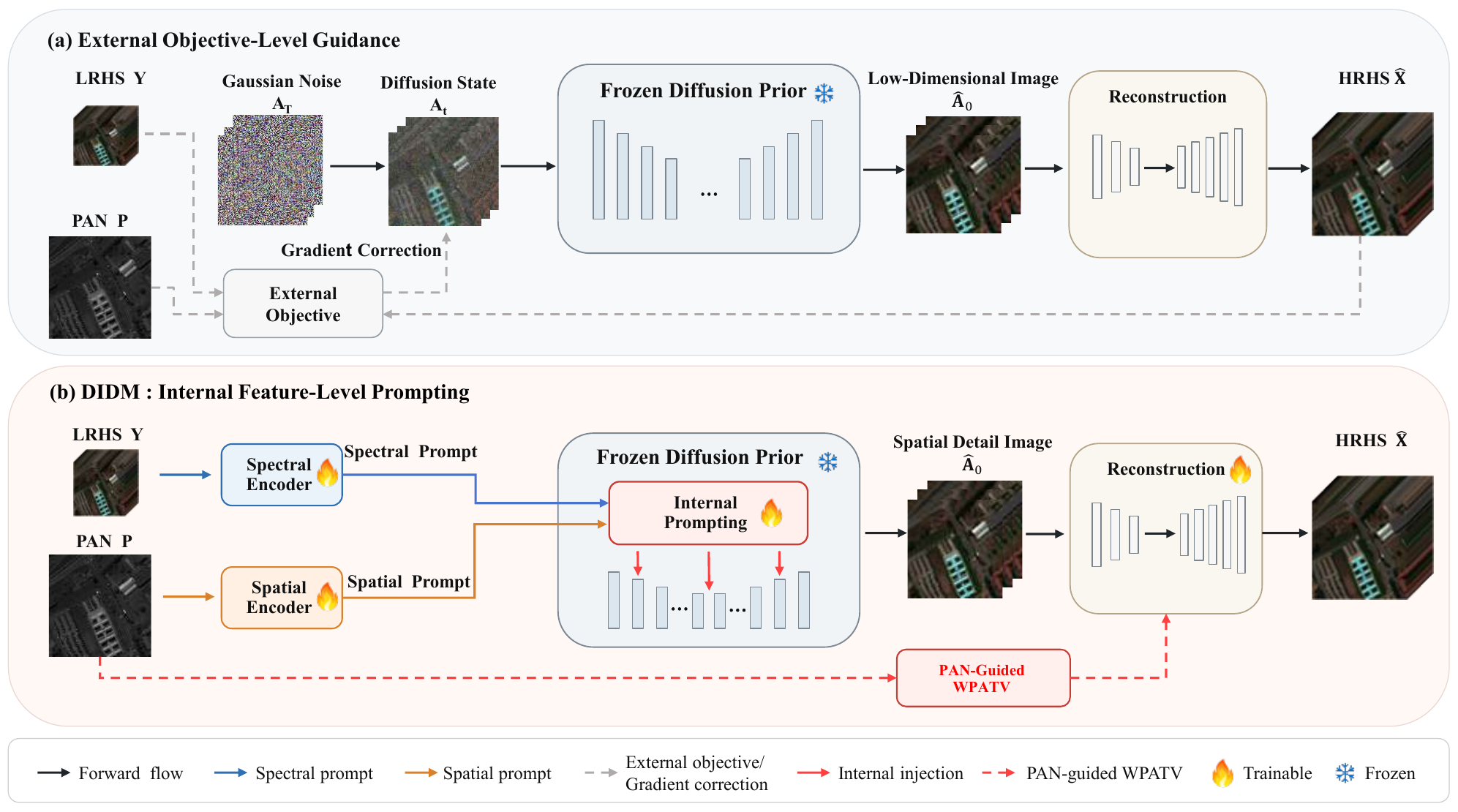}
    \caption{Conceptual comparison between external objective-level guidance and the proposed internal feature-level prompting. The upper panel summarizes the common mechanism of PLRDiff, HIR-Diff, and DM-ZS under a unified LRHS/PAN reconstruction setting: observations and auxiliary priors define an external differentiable objective on the current HRHS estimate, whose gradient corrects the current diffusion state $\mathbf{A}_{t}$ while the pre-trained diffusion prior remains frozen. The lower panel presents DIDM, which retains the standard reverse-diffusion process but encodes LRHS and PAN into spectral and spatial prompts, respectively, allowing both observations to participate directly in the evolution of internal diffusion features. The resulting spatial detail image $\widehat{\mathbf{A}}_{0}$ is used to reconstruct HRHS, while the PAN-guided WPATV imposes structural regularization on the reconstruction process. To emphasize the difference between the two conditioning mechanisms, the LRHS degradation fidelity and the PAN response fidelity retained by DIDM are not explicitly shown.}
    \label{fig:guide_map}
\end{figure*}

However, the success of this common pipeline does not resolve how task observations should interact with the frozen diffusion prior. As illustrated in Fig.~\ref{fig:guide_map}(a), representative methods construct observation-fidelity terms and auxiliary priors outside the diffusion backbone. At each reverse step, these terms are evaluated on the reconstructed HRHS candidate, and their gradient is propagated back to correct the current diffusion state $\mathbf{A}_{t}$. This mechanism effectively steers the sampling trajectory, but LRHS and PAN do not enter the pre-trained U-Net as condition tokens or feature-level inputs. The first unresolved problem is therefore the conditioning interface: how can the complementary spectral evidence of LRHS and spatial evidence of PAN participate directly in the internal feature evolution of the frozen prior, rather than only correcting its output trajectory through an external objective?

A second problem concerns the use of PAN structure during zero-shot reconstruction. LRHS degradation fidelity maintains spectral consistency under the spatial degradation model, and PAN response fidelity enforces consistency with the PAN observation. Both terms are indispensable and are retained in our formulation. However, they constrain the reconstructed image mainly in the observation space and do not explicitly distinguish boundaries from homogeneous regions when regularizing the spatial component. PAN gradients provide a physically meaningful cue for this distinction: object boundaries, road edges, and texture transitions indicate where spatial variations should be preserved, whereas homogeneous regions indicate where unnecessary fluctuations should be suppressed. The reconstruction objective should therefore complement observation fidelity with a PAN-aware structural prior.

To address these limitations, we propose dual-modality image-prompted diffusion model, termed DIDM, for hyperspectral pansharpening. As shown in Fig.~\ref{fig:guide_map}(b), DIDM retains the standard reverse-diffusion process and a frozen remote-sensing diffusion prior while redesigning how the observations interact with the backbone. LRHS and PAN are encoded into spectral and spatial condition tokens, respectively, and injected into intermediate diffusion features through cross-attention \cite{Vaswani2017Attention}. In this way, LRHS-derived spectral evidence and PAN-derived spatial structures participate in reverse sampling as internal feature-level prompts, producing a prompt-conditioned spatial detail image $\widehat{\mathbf{A}}_{0}$ that is mapped to HRHS by the NSSD reconstruction backend inherited from DM-ZS. DIDM further introduces a PAN-guided weighted pixel-aware total variation (WPATV) term adapted from XINet \cite{Li2023XIA}. PAN-gradient-derived inverse weights impose edge-aware regularization on the spatial component, relaxing smoothing near PAN-aligned structures while suppressing spurious variations in homogeneous regions. The resulting objective jointly enforces spectral fidelity, PAN response consistency, and local structural regularity.

Our main contributions are as follows:
\begin{itemize}
    \item[\textcolor{black}{$\bullet$}] We propose DIDM, a dual-modality image-prompted diffusion model for zero-shot hyperspectral pansharpening. It combines a frozen remote-sensing diffusion prior, modality-specific image prompts, and an NSSD reconstruction backend without requiring external paired HRHS samples.
    \item[\textcolor{black}{$\bullet$}] We establish an internal feature-level conditioning interface for the frozen diffusion prior. LRHS-derived spectral prompts and PAN-derived spatial prompts interact with intermediate diffusion features during reverse sampling, in contrast to representative methods that primarily steer the current diffusion state through an external differentiable objective.
    \item[\textcolor{black}{$\bullet$}] We introduce a PAN-guided spatial--spectral--structural objective. In addition to LRHS degradation fidelity and PAN response fidelity, WPATV imposes edge-aware regularization on the spatial component, preserving PAN-aligned discontinuities while suppressing fluctuations in homogeneous regions.
    \item[\textcolor{black}{$\bullet$}] We provide a comprehensive evaluation under complementary reduced-resolution and full-resolution settings. Results on Pavia, Chikusei, and Houston demonstrate superior reference-based accuracy, while the FR1 sample shows favorable observation consistency. Ablation, efficiency, and diagnostic analyses further validate the dual-modality prompting, the PAN-guided structural regularization, and the multi-stage prompt injection.
\end{itemize}

\section{Related Work}
\label{sec:rw}

\subsection{Hyperspectral Pansharpening}

Hyperspectral pansharpening has been extensively studied under model based and learning based paradigms~\cite{Loncan2015Review,Ciotola2025CriticalReview}. Model based methods reconstruct high resolution hyperspectral images by combining explicit observation models with handcrafted spatial or spectral priors. GSA estimates the relationship between spectral bands and PAN through regression~\cite{Aiazzi2007GSA}, while MTF tailored multiscale approaches inject spatial details using sensor aware filtering~\cite{Aiazzi2006MTFGLP}. CNMF couples hyperspectral and auxiliary observations through nonnegative matrix factorization~\cite{Yokoya2012CNMF}, and HySure formulates hyperspectral super resolution through convex subspace regularization~\cite{Simoes2015HySure}. Sparse representation and Bayesian fusion exploit low dimensional spectral structures~\cite{Akhtar2014Sparse,Wei2015BayesianFusion}, whereas TV based methods impose spatial or spectral gradient regularity to preserve boundaries and suppress artifacts~\cite{Addesso2017CTV,Takeyama2020HSSTV}. These methods are interpretable and require no external training data, but their manually specified priors may limit their ability to model complex spatial and spectral structures.

Learning-based approaches improve representation capacity through data-driven joint spatial--spectral modeling. CNN and residual architectures learn spatial-detail injection from simulated training pairs~\cite{Masi2016PNN,Yang2017PanNet,Wei2017DRPNN,Yuan2018MSDCNN}, while deep hyperspectral image fusion networks incorporate deep priors, spatial--spectral attention, external--internal attention, and U-shaped architectures to improve spectral preservation and spatial reconstruction~\cite{Han2020DHPDARN,Hu2022SSANet,Liu2022SSAU}. Deep-unfolding and degradation-aware methods further improve model interpretability by integrating observation models into network optimization~\cite{Dian2018DHSIS,Guo2022EIA,Xie2022MHFNet}. Recent studies also explore selective re-learning of degraded spatial--spectral features~\cite{Liu2025SRLFNet} and cross-scale designs for generalization across spatial resolutions~\cite{Cao2026ScaleFormer}. To reduce dependence on paired HRHS supervision, zero-shot methods directly optimize task-specific reconstruction from the observed LRHS and PAN pair. ZSL performs zero-shot hyperspectral super-resolution from the test observations~\cite{Qu2022ZSL}, $\rho$-PNN introduces band-wise adaptation with hysteresis-based spectral-quality control~\cite{Guarino2025RhoPNN}, and Hipandas jointly addresses denoising and pansharpening through guided reconstruction networks and low-rank priors~\cite{Xu2025Hipandas}. These methods avoid external paired training by adapting the reconstruction process directly to the target observations. In contrast, DIDM retains per-image optimization while exploiting a pre-trained remote-sensing diffusion prior through internal dual-modality conditioning and PAN-guided structural regularization.

\subsection{Diffusion Priors and Conditioning Interfaces}

Diffusion models serve as powerful generative priors for image synthesis and inverse problems. DDPM and DDIM establish the fundamental iterative denoising process~\cite{Ho2020DDPM,Song2021DDIM}, while subsequent studies further improve generation quality and conditional sampling~\cite{Nichol2021ImprovedDDPM,Dhariwal2021DiffusionBeatGANs}. Diffusion priors are also widely adopted for various image restoration tasks. SR3 performs image super-resolution through iterative refinement, RePaint applies diffusion sampling to image inpainting, and LDM transfers generation into a compact latent space~\cite{Saharia2023SR3,Lugmayr2022RePaint,Rombach2022LDM}. DDRM and DPS further incorporate observation models into diffusion-based inverse reconstruction~\cite{Kawar2022DDRM,Chung2023DPS}. This capability is particularly attractive for hyperspectral pansharpening applications, where paired full-resolution supervision is difficult to acquire.

Existing hyperspectral diffusion methods mainly adapt low-dimensional image priors through reverse sampling followed by spatial--spectral reconstruction. HIR-Diff combines diffusion sampling with a low-rank spatial--spectral representation for unsupervised hyperspectral restoration, PLRDiff extends low-rank diffusion modeling to hyperspectral pansharpening~\cite{Rui2024PLRDiff}, and DM-ZS combines iterative zero-shot guidance with NSSD reconstruction for each LRHS/PAN pair~\cite{Xiao2025DMZS}. Although their objectives and reconstruction operators differ, these approaches mainly enforce observation consistency through external differentiable objectives whose gradients correct the current diffusion state. The observations therefore remain weakly coupled with the internal diffusion features. DIDM complements this paradigm by retaining image-level fidelity for reconstruction supervision while introducing LRHS/PAN information as internal feature-level prompts.

A related line of research studies how explicit conditions control pretrained diffusion models. Textual Inversion and DreamBooth introduce concept-specific conditioning through learned embeddings or parameter adaptation~\cite{Gal2023TextualInversion,Ruiz2023DreamBooth}, while GLIGEN, ControlNet, and T2I-Adapter inject external visual conditions through trainable modules~\cite{Li2023GLIGEN,Zhang2023ControlNet,Mou2024T2IAdapter}. Recent conditional diffusion frameworks further demonstrate the benefit of structured visual and motion priors for controllable generation, including pose-guided person synthesis~\cite{shen2024imagpose}, customizable virtual dressing~\cite{shen2025imagdressing}, fine-grained garment generation~\cite{shen2026imaggarment}, and long-term talking-face generation with motion priors~\cite{shenlong}. However, these methods mainly target natural-image or video synthesis and rely on semantic, geometric, or motion conditions. Hyperspectral pansharpening instead involves two physically coupled observations carrying complementary spectral and spatial information. DIDM therefore develops a task-specific conditioning interface in which LRHS provides spectral prompts and PAN provides spatial prompts through cross-attention, while PAN-derived structural regularization further constrains zero-shot reconstruction.

\section{Proposed Method}\label{sec:method}

\subsection{Problem Formulation}

Let $\mathbf{Y}\in\mathbb{R}^{B\times h\times w}$ denote the observed LRHS with $B$ spectral bands, where $h$ and $w$ are its spatial height and width, respectively, and let $\mathbf{P}\in\mathbb{R}^{1\times H\times W}$ denote the observed PAN image, where $H$ and $W$ are the corresponding high-resolution spatial height and width. Let $\mathbf{X}\in\mathbb{R}^{B\times H\times W}$ denote the latent HRHS, and let $\widehat{\mathbf{X}}\in\mathbb{R}^{B\times H\times W}$ denote the reconstructed HRHS. Here, $H=sh$ and $W=sw$, where $s$ is the spatial scale factor. Following the common observation model,
\begin{equation}
    \mathbf{Y}\approx\mathcal{D}(\mathbf{X}),\quad \mathbf{P}\approx\mathcal{R}(\mathbf{X}),
    \label{eq:observation_model}
\end{equation}
where $\mathcal{D}(\cdot)$ denotes spatial degradation, and $\mathcal{R}(\cdot)$ denotes the spectral response from HRHS to PAN.

DIDM retains the standard reverse-diffusion process and uses a frozen remote-sensing pre-trained diffusion backbone as the spatial detail generation prior. The method consists of three stages. First, LRHS and PAN are encoded into spectral and spatial tokens, respectively, and introduced into intermediate diffusion features through Dual-Prompt Cross-Attention. Second, the prompt-conditioned diffusion backbone estimates a three-channel spatial detail image $\widehat{\mathbf{A}}_{0}\in\mathbb{R}^{3\times H\times W}$, hereafter referred to as the spatial detail image. Third, the spatial detail image and LRHS are transformed into HRHS through an NSSD-based spatial--spectral reconstruction backend. The optimization retains the LRHS degradation fidelity and the PAN response fidelity and further incorporates the PAN-guided WPATV on the spatial branch. In this way, the method combines internal feature-level prompting with a spatial--spectral--structural reconstruction objective while keeping the original pre-trained diffusion parameters fixed.

The reconstruction backend follows the formulation adopted in DM-ZS so that the contribution of the conditioning interface can be isolated from changes in the subsequent HRHS reconstruction. Unlike external objective-level guidance, the proposed interface allows LRHS-derived spectral evidence and PAN-derived spatial evidence to participate directly in the evolution of intermediate diffusion features. The following subsections describe the dual-modality prompt injection, the prompt-conditioned detail generation, the spatial--spectral reconstruction, and the optimization objective.

\subsection{Dual-Modality Image Prompt Injection}

DIDM converts LRHS and PAN into modality-specific condition tokens. LRHS provides spectral cues, whereas PAN provides high-resolution spatial structures. The token generation is formulated as
\begin{equation}
    \mathbf{T}_{\mathrm{spe}}=\mathcal{E}_{\mathrm{spe}}(\mathbf{Y}),\quad
    \mathbf{T}_{\mathrm{spa}}=\mathcal{E}_{\mathrm{spa}}(\mathbf{P}),
    \label{eq:prompt_encoder}
\end{equation}
where $\mathcal{E}_{\mathrm{spe}}(\cdot)$ and $\mathcal{E}_{\mathrm{spa}}(\cdot)$ denote the spectral and spatial prompt encoders, respectively. The resulting tokens are $\mathbf{T}_{\mathrm{spe}}\in\mathbb{R}^{N_s\times C}$ and $\mathbf{T}_{\mathrm{spa}}\in\mathbb{R}^{N_p\times C}$, where $N_s$ and $N_p$ are the numbers of spectral and spatial tokens, and $C$ is the token dimension. We further denote by $\alpha\in[0,1]$ the spectral-token ratio controlling their relative spectral--spatial composition; $\alpha=0$ and $\alpha=1$ correspond to spatial-only and spectral-only prompting, respectively, while intermediate values retain both modalities at different proportions. Both encoders are lightweight two-layer convolutional networks. This design projects the two observations into a common condition space without excessive learnable capacity, which is important for per-image zero-shot optimization.

\begin{figure}[t]
    \centering
    \includegraphics[width=\columnwidth]{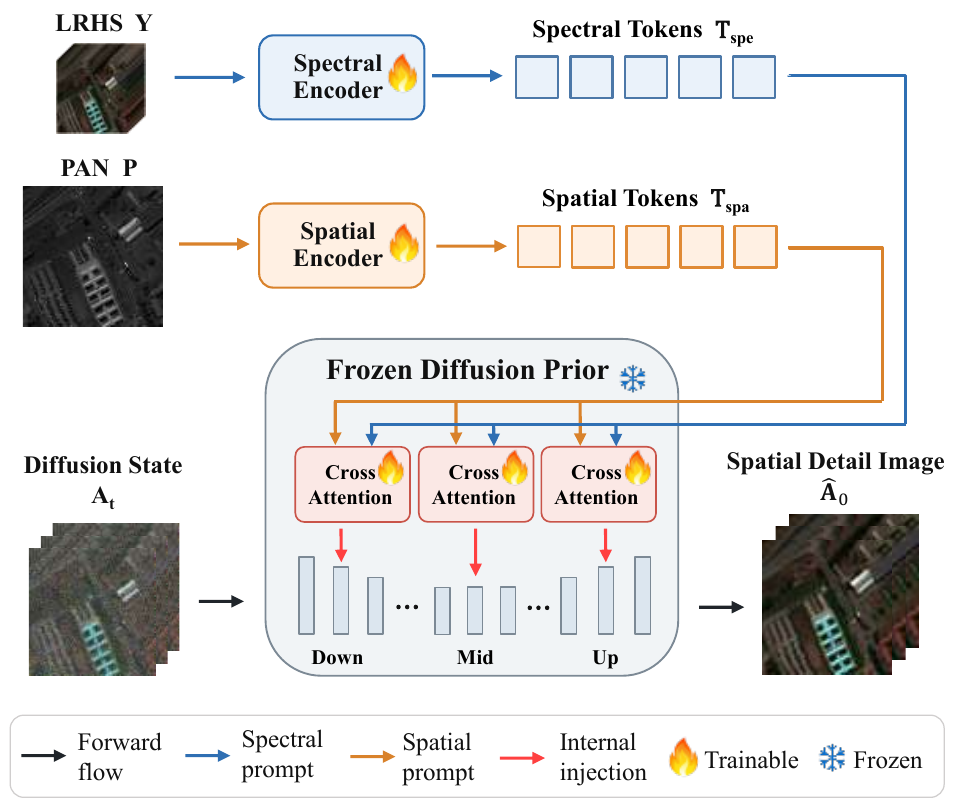}
    \caption{Dual-modality prompt injection in DIDM. LRHS and PAN are encoded into spectral and spatial tokens, respectively. At the representative Down, Mid, and Up stages, local diffusion features provide the queries, while the two token groups provide modality-specific keys and values to trainable Cross Attention blocks. The resulting responses are injected into the corresponding frozen diffusion features, and the prompt-conditioned diffusion prior estimates the spatial detail image $\widehat{\mathbf{A}}_{0}$.}
    \label{fig:cross_attention}
\end{figure}

Fig.~\ref{fig:cross_attention} illustrates the prompt injection process. The same spectral and spatial token groups are supplied to representative Cross Attention blocks in the Down, Mid, and Up stages of the frozen diffusion prior. Each illustrated block receives both modalities, and its output is injected into the corresponding diffusion features. The figure abstracts the internal attention computation to maintain readability at single-column width; the complete query, key, value, attention response, and residual feature update are defined below.

Let $\mathbf{F}_{t}^{l}\in\mathbb{R}^{N_l\times C_l}$ denote the diffusion feature at reverse step $t$ and injection layer $l$, where $N_l$ is the number of flattened spatial locations and $C_l$ is the feature dimension at that layer. For $m\in\{\mathrm{spe},\mathrm{spa}\}$, which indexes the spectral or spatial prompt branch, the prompt-injection computation is given by
\begin{equation}
    \mathbf{Q}_{t}^{l}=\mathbf{F}_{t}^{l}\mathbf{W}_{Q}^{l},\quad
    \mathbf{K}_{m}^{l}=\mathbf{T}_{m}\mathbf{W}_{K,m}^{l},\quad
    \mathbf{V}_{m}^{l}=\mathbf{T}_{m}\mathbf{W}_{V,m}^{l},
    \label{eq:qkv_definition}
\end{equation}
\begin{equation}
    \mathbf{C}_{m,t}^{l}
    =
    \mathrm{Attn}(\mathbf{Q}_{t}^{l},\mathbf{K}_{m}^{l},\mathbf{V}_{m}^{l}),
    \label{eq:compact_cross_attention}
\end{equation}
\begin{equation}
    \mathrm{Attn}(\mathbf{Q},\mathbf{K},\mathbf{V})
    =
    \mathrm{softmax}(\mathbf{Q}\mathbf{K}^{\top}/\sqrt{d_l})\mathbf{V},
    \label{eq:attention_operator}
\end{equation}
\begin{equation}
    \widetilde{\mathbf{F}}_{t}^{l}
    =
    \mathrm{SA}(\mathbf{F}_{t}^{l})
    +
    \sum_{m\in\{\mathrm{spe},\mathrm{spa}\}}
    \eta_{m}^{l}\mathbf{C}_{m,t}^{l},
    \label{eq:prompt_fusion}
\end{equation}
where $\mathbf{Q}_{t}^{l}$ is the query projected from the diffusion feature, $\mathbf{K}_{m}^{l}$ and $\mathbf{V}_{m}^{l}$ are the key and value projected from modality-$m$ tokens, and $\mathbf{C}_{m,t}^{l}$ is the corresponding cross-attention response. The matrices $\mathbf{W}_{Q}^{l}$, $\mathbf{W}_{K,m}^{l}$, and $\mathbf{W}_{V,m}^{l}$ denote the query, key, and value projections, respectively, and $d_l$ is the attention dimension. Moreover, $\mathrm{SA}(\cdot)$ denotes the original self-attention operation, $\eta_{m}^{l}$ is the modality-balancing coefficient at layer $l$, and $\widetilde{\mathbf{F}}_{t}^{l}\in\mathbb{R}^{N_l\times C_l}$ is the resulting prompt-conditioned feature. Thus, the diffusion feature supplies the query, while the LRHS- and PAN-derived tokens provide complementary conditioning contexts that participate in reverse diffusion without modifying the pre-trained backbone.

\subsection{Prompt-Conditioned Detail Generation and Spatial--Spectral Reconstruction}

Let $\mathbf{A}_{0}\in\mathbb{R}^{3\times H\times W}$ denote the spatial detail image modeled by the diffusion prior, and let $\mathbf{A}_{t}\in\mathbb{R}^{3\times H\times W}$ denote its noisy state at timestep $t$. Conditioned on the two token groups, the denoising network predicts the noise and derives the clean estimate as
\begin{equation}
    \boldsymbol{\epsilon}_{t}
    =
    \boldsymbol{\epsilon}_{\theta}
    (\mathbf{A}_{t},t,\mathbf{T}_{\mathrm{spe}},\mathbf{T}_{\mathrm{spa}}),
    \label{eq:noise_prediction}
\end{equation}
\begin{equation}
    \widehat{\mathbf{A}}_{0}^{t}
    =
    \frac{
    \mathbf{A}_{t}
    -
    \sqrt{1-\bar{\gamma}_{t}}\,
    \boldsymbol{\epsilon}_{t}}
    {\sqrt{\bar{\gamma}_{t}}},
    \label{eq:a0_estimation}
\end{equation}
where $\boldsymbol{\epsilon}_{\theta}(\cdot)$ is the denoising network with fixed pre-trained parameters $\theta$, $\boldsymbol{\epsilon}_{t}\in\mathbb{R}^{3\times H\times W}$ is the predicted noise, $\bar{\gamma}_{t}$ is the cumulative noise-scheduling coefficient, and $\widehat{\mathbf{A}}_{0}^{t}\in\mathbb{R}^{3\times H\times W}$ is the clean spatial-detail estimate obtained at step $t$. The final estimate after reverse diffusion is denoted by $\widehat{\mathbf{A}}_{0}$.

The spatial detail image provides a compact high-resolution structural representation, whereas the target HRHS contains $B$ spectral bands. We therefore retain the NSSD-based reconstruction backend from DM-ZS to map the spatial detail image and LRHS spectral information to the HRHS space. Following DM-ZS \cite{Xiao2025DMZS}, the NSSD reconstruction adopts a low-dimensional subspace with rank $r$. Let $\mathcal{M}_{\phi}(\cdot)$ denote the complete reconstruction operator with trainable parameters $\phi$. It contains a U-shaped spatial branch and a spectral mapping branch and reconstructs HRHS as
\begin{equation}
    \widehat{\mathbf{X}}
    =
    \mathcal{M}_{\phi}(\widehat{\mathbf{A}}_{0},\mathbf{Y}),
    \label{eq:reconstruction}
\end{equation}
while the output of the spatial branch used for the PAN-guided structural regularization is denoted by
\begin{equation}
    \widehat{\mathbf{S}}
    =
    \mathcal{S}_{\phi}(\widehat{\mathbf{A}}_{0}),
    \label{eq:spatial_component}
\end{equation}
where $\mathcal{S}_{\phi}(\cdot)$ is the spatial branch of $\mathcal{M}_{\phi}(\cdot)$, $\widehat{\mathbf{S}}\in\mathbb{R}^{C_s\times H\times W}$ is its output, and $C_s$ is the number of spatial-feature channels. The single-channel PAN-derived weight defined below is shared across these channels. Retaining this backend preserves a clear connection with prior diffusion-based hyperspectral pansharpening and allows DIDM to focus on improving the preceding observation-to-diffusion conditioning interface.

\subsection{PAN-Guided Spatial--Spectral--Structural Objective}

\begin{figure}[t]
    \centering
    \includegraphics[width=\columnwidth]{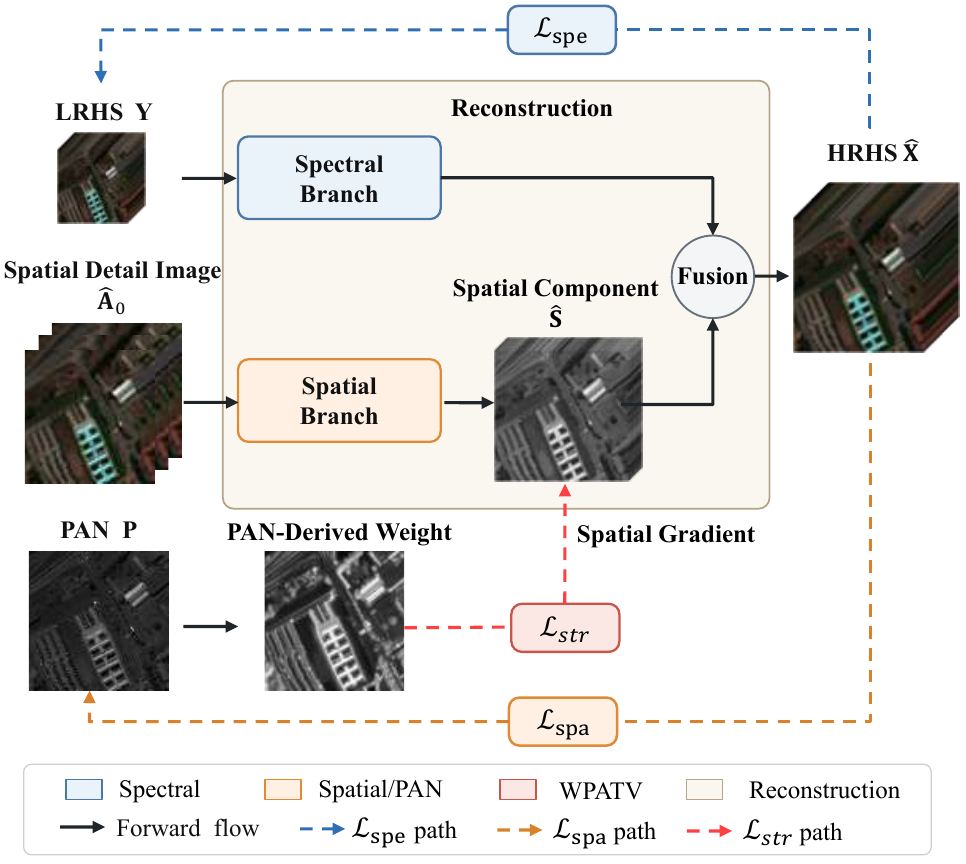}
    \caption{Spatial--spectral reconstruction and objective attachment in DIDM. The spatial detail image $\widehat{\mathbf{A}}_{0}$ and LRHS $\mathbf{Y}$ are processed by the spatial and spectral branches, respectively, and their representations are fused to reconstruct HRHS $\widehat{\mathbf{X}}$. The blue and orange dashed paths denote LRHS degradation fidelity $\mathcal{L}_{\mathrm{spe}}$ and PAN response fidelity $\mathcal{L}_{\mathrm{spa}}$, respectively. The single-channel PAN-derived weight $\mathbf{W}$ modulates the gradients of the multi-channel spatial component $\widehat{\mathbf{S}}$, yielding the PAN-guided WPATV term $\mathcal{L}_{\mathrm{str}}$.}
    \label{fig:wpatv}
\end{figure}

The objective contains three complementary terms, whose connections to the reconstruction process are illustrated in Fig.~\ref{fig:wpatv}. The spatial detail image and LRHS are processed by the spatial and spectral branches, respectively, and their representations are fused to reconstruct $\widehat{\mathbf{X}}$. The blue and orange dashed paths denote LRHS degradation fidelity and PAN response fidelity, whereas the red path indicates that PAN-guided WPATV acts on the spatial gradients of $\widehat{\mathbf{S}}$. For clarity, the figure abstracts the degradation operator, response operator, and construction of the PAN-derived weight, which are defined below.

The LRHS degradation fidelity and the PAN response fidelity are respectively formulated as
\begin{equation}
    \mathcal{L}_{\mathrm{spe}}
    =
    \|\mathcal{D}(\widehat{\mathbf{X}})-\mathbf{Y}\|_{1},
    \label{eq:spectral_loss}
\end{equation}
\begin{equation}
    \mathcal{L}_{\mathrm{spa}}
    =
    \|\mathcal{R}(\widehat{\mathbf{X}})-\mathbf{P}\|_{1},
    \label{eq:spatial_loss}
\end{equation}
where $\mathcal{D}(\cdot)$ and $\mathcal{R}(\cdot)$ are defined in Eq.~\eqref{eq:observation_model}, and $\|\cdot\|_{1}$ denotes the sum of absolute residual values. The terms $\mathcal{L}_{\mathrm{spe}}$ and $\mathcal{L}_{\mathrm{spa}}$ therefore measure LRHS degradation fidelity and PAN response fidelity, respectively.

Although these two fidelity terms are necessary for observation consistency, they mainly constrain the reconstructed HRHS at the image level and do not explicitly impose a pixel-wise edge-aware prior on the spatial branch. PAN contains high-resolution edges, object boundaries, and texture transitions, which indicate where spatial variations should be retained. We therefore adapt the WPATV principle to construct a PAN-guided structural regularizer.

The PAN gradient map, clipped gradient map, pixel-aware weight, and structural regularization term are defined as
\begin{equation}
    \mathbf{G}
    =
    \sum_{c}
    (|\nabla_x\mathbf{P}_{c}|+|\nabla_y\mathbf{P}_{c}|),
    \quad
    \bar{\mathbf{G}}
    =
    \min(\mathbf{G},Q_p(\mathbf{G})),
    \label{eq:wpatv_gradient_clip}
\end{equation}
\begin{equation}
    \mathbf{W}=g(|\nabla\mathbf{P}|),
    \label{eq:wpatv_weight_mapping}
\end{equation}
\begin{equation}
    \mathbf{W}
    =
    1-
    \frac{
    \bar{\mathbf{G}}-g_{\min}}
    {g_{\max}-g_{\min}+\epsilon},
    \label{eq:wpatv_weight}
\end{equation}
\begin{equation}
    \mathcal{L}_{\mathrm{str}}
    =
    \mathrm{mean}
    (\mathbf{W}\odot|\nabla_x\widehat{\mathbf{S}}|)
    +
    \mathrm{mean}
    (\mathbf{W}\odot|\nabla_y\widehat{\mathbf{S}}|),
    \label{eq:wpatv_loss}
\end{equation}
where $c$ indexes the PAN channels and $\mathbf{P}_{c}$ is the $c$th channel; for the single-channel PAN used here, the summation contains one term. The operators $\nabla_x$ and $\nabla_y$ denote horizontal and vertical finite differences, respectively, and $|\cdot|$ denotes element-wise absolute value. Accordingly, $\mathbf{G}\in\mathbb{R}^{1\times H\times W}$ is the anisotropic PAN gradient-response map. Here, $p\in(0,1)$ denotes the winsorization percentile expressed as a quantile level, $Q_p(\mathbf{G})$ gives the corresponding clipping threshold, and $\bar{\mathbf{G}}$ is obtained by element-wise clipping. In Eq.~\eqref{eq:wpatv_weight_mapping}, $|\nabla\mathbf{P}|$ is shorthand for $\mathbf{G}$, and $g(\cdot)$ denotes the normalized inverse-gradient mapping defined in Eq.~\eqref{eq:wpatv_weight}. Here, $g_{\min}$ and $g_{\max}$ are the minimum and maximum of $\bar{\mathbf{G}}$, $\epsilon=10^{-8}$ is a fixed constant for numerical stability, and $\mathbf{W}\in\mathbb{R}^{1\times H\times W}$ is the resulting weight map. In Eq.~\eqref{eq:wpatv_loss}, $\odot$ denotes element-wise multiplication, $\mathrm{mean}(\cdot)$ averages all entries, and $\mathcal{L}_{\mathrm{str}}$ is the PAN-guided WPATV loss. The finite-difference maps retain the original spatial size, and $\mathbf{W}$ is broadcast across the $C_s$ channels of $\widehat{\mathbf{S}}$. Thus, smoothing is relaxed near strong PAN gradients and strengthened in homogeneous regions.

The overall objective is
\begin{equation}
    \mathcal{L}
    =
    \lambda_{\mathrm{spe}}\mathcal{L}_{\mathrm{spe}}
    +
    \lambda_{\mathrm{spa}}\mathcal{L}_{\mathrm{spa}}
    +
    \lambda_{\mathrm{str}}\mathcal{L}_{\mathrm{str}},
    \label{eq:overall_loss}
\end{equation}
where $\mathcal{L}$ is the overall objective, and $\lambda_{\mathrm{spe}}$, $\lambda_{\mathrm{spa}}$, and $\lambda_{\mathrm{str}}$ balance the three loss terms. This spatial--spectral--structural objective extends observation fidelity with PAN-guided local regularization, thereby encouraging spectral consistency, PAN response consistency, and PAN-aligned structural preservation in the reconstructed HRHS.

\subsection{Optimization Procedure}

Algorithm~\ref{alg:dmip} summarizes the zero-shot optimization procedure. For each LRHS/PAN pair, the pre-trained diffusion parameters $\theta$ remain fixed during per-pair adaptation. Let $\Theta=\{\theta_{\mathrm{spe}},\theta_{\mathrm{spa}},\theta_{\mathrm{inj}},\phi\}$ collect all learnable parameters, where $\theta_{\mathrm{spe}}$ and $\theta_{\mathrm{spa}}$ parameterize the two prompt encoders and $\theta_{\mathrm{inj}}$ contains the parameters of the introduced prompt-injection modules, including the coefficients $\eta_m^l$. Thus, $\eta_m^l$ is optimized as part of the model rather than supplied as an external input. The fixed settings required by the procedure include the reverse-diffusion schedule, the spectral-token ratio $\alpha$, the reconstruction rank $r$, the WPATV parameters $p$, the three loss weights, and the optimization stopping criterion. The operators $Q_p(\cdot)$ and $g(\cdot)$ are deterministic mappings defined in Eqs.~\eqref{eq:wpatv_gradient_clip}--\eqref{eq:wpatv_weight}; once $p$ and $\epsilon$ are specified, they introduce no additional free input.

\begin{algorithm}[t]
\caption{Optimization procedure of DIDM}
\label{alg:dmip}
\begin{algorithmic}[1]
\REQUIRE LRHS $\mathbf{Y}$; PAN $\mathbf{P}$; operators $\mathcal{D}$ and $\mathcal{R}$; fixed diffusion backbone $\boldsymbol{\epsilon}_{\theta}$ and reverse-diffusion schedule; spectral-token ratio $\alpha$; reconstruction rank $r$; WPATV parameters $p$; loss weights $\lambda_{\mathrm{spe}}$, $\lambda_{\mathrm{spa}}$, and $\lambda_{\mathrm{str}}$; maximum optimization iterations $K$.
\STATE Initialize the learnable parameter set $\Theta$ under the prescribed $\alpha$ and $r$; keep $\theta$ fixed.
\STATE Construct the PAN-guided weight $\mathbf{W}$ from $\mathbf{P}$ using Eqs.~\eqref{eq:wpatv_gradient_clip}--\eqref{eq:wpatv_weight}.
\FOR{$k=1$ to $K$}
    \STATE Generate $\mathbf{T}_{\mathrm{spe}}$ and $\mathbf{T}_{\mathrm{spa}}$ using Eq.~\eqref{eq:prompt_encoder}.
    \STATE Perform prompt-conditioned reverse diffusion, applying Eqs.~\eqref{eq:qkv_definition}--\eqref{eq:prompt_fusion} at the injected layers, to obtain $\widehat{\mathbf{A}}_{0}$.
    \STATE Obtain $\widehat{\mathbf{X}}$ and $\widehat{\mathbf{S}}$ using Eqs.~\eqref{eq:reconstruction} and \eqref{eq:spatial_component}.
    \STATE Compute $\mathcal{L}_{\mathrm{spe}}$, $\mathcal{L}_{\mathrm{spa}}$, and $\mathcal{L}_{\mathrm{str}}$ using Eqs.~\eqref{eq:spectral_loss}, \eqref{eq:spatial_loss}, and \eqref{eq:wpatv_loss}, respectively.
    \STATE Update $\Theta$ by minimizing Eq.~\eqref{eq:overall_loss}; keep $\theta$ fixed.
\ENDFOR
\RETURN Reconstructed HRHS $\widehat{\mathbf{X}}$.
\end{algorithmic}
\end{algorithm}

\section{Experiments and Analysis}\label{sec:exp}

\subsection{Experimental Settings}

\noindent\textbf{Datasets and Protocols.} We evaluate DIDM under two complementary protocols commonly adopted for hyperspectral pansharpening: reduced-resolution (RR) reference-based assessment and full-resolution (FR) no-reference assessment on real multiresolution observations. The RR experiments are conducted on three public hyperspectral datasets: Pavia,\footnote{\url{https://rslab.ut.ac.ir/data}} Chikusei,\footnote{\url{https://naotoyokoya.com/Download.html}} and Houston.\footnote{\url{https://hyperspectral.ee.uh.edu/?page_id=459}} Following Wald's protocol \cite{Wald2000Quality}, each original hyperspectral crop is treated as the reference HRHS image, and the corresponding LRHS observation is generated by applying a $9\times9$ Gaussian blur to each spectral band, followed by fourfold bicubic downsampling. Because paired PAN observations are unavailable, the PAN image is simulated by uniformly averaging the spectral bands within the available visible and near-infrared (VIS--NIR) range; bands 16--81 under one-based indexing are used for Chikusei, whereas all retained bands are used for Pavia and Houston. The FR assessment is performed on the released FR1 full-resolution sample from HyperPanCollection\footnote{\url{https://github.com/liangjiandeng/HyperPanCollection}} \cite{vivone2022panchromatic}, using its real PAN and hyperspectral observations at their native resolutions without constructing a reference HRHS image.

\textbf{\emph{(1) Pavia}} is an urban hyperspectral dataset acquired by the Reflective Optics System Imaging Spectrometer. We use eight $256\times256$ crops with 93 spectral bands. The scenes contain representative urban materials and structures, including buildings, roads, shadows, and vegetation, and are therefore used to assess whether PAN-aligned geometric details, such as roof contours and road boundaries, can be recovered without compromising spectral fidelity.

\textbf{\emph{(2) Chikusei}} was acquired over mixed agricultural and urban areas in Chikusei, Ibaraki, Japan. We use ten cropped samples of size $256\times256$, each comprising 128 spectral bands. Its fine field boundaries, agricultural textures, and narrow linear structures provide a demanding test of local-detail reconstruction and high-dimensional spectral consistency across heterogeneous land-cover regions.

\textbf{\emph{(3) Houston}} is provided by the Hyperspectral Image Analysis Laboratory at the University of Houston. We use ten cropped samples of size $256\times256$ with 144 spectral bands. The dataset contains dense man-made structures, complex object boundaries, and pronounced spectral variability, thereby challenging both spatial reconstruction and spectral preservation. Owing to this complexity and its high spectral dimensionality, Houston is also adopted as the representative dataset for the ablation and sensitivity studies.

\textbf{\emph{(4) FR1}} provides a real multiresolution PAN--hyperspectral observation pair for complementary FR assessment. The released sample used in our experiments contains a PAN image of size $240\times240$ and a corresponding LRHS observation of size $40\times40$ with 69 spectral bands, resulting in a spatial-resolution ratio of six. Because a ground-truth HRHS image is unavailable, FR1 is used solely to examine no-reference observation consistency and visual behavior at the native resolution. All datasets used in this study are publicly available through the links provided above.

\noindent\textbf{Compared Methods.}
We compare DIDM with nine representative competing methods spanning three major methodological families commonly considered in hyperspectral pansharpening. The classical model-based methods include GSA \cite{Aiazzi2007GSA}, CNMF \cite{Yokoya2012CNMF}, and TV \cite{Addesso2017CTV}, representing component-substitution, matrix-factorization, and variational-regularization approaches, respectively. The direct zero-shot reconstruction methods include ZSL \cite{Qu2022ZSL}, $\rho$-PNN \cite{Guarino2025RhoPNN}, and Hipandas \cite{Xu2025Hipandas}, which adapt task-specific reconstruction directly from the available observations without external paired supervision. The diffusion-prior methods comprise PLRDiff \cite{Rui2024PLRDiff}, HIR-Diff \cite{Pang2024HIRDiff}, and DM-ZS \cite{Xiao2025DMZS}. This selection provides broad coverage of established model-based approaches, recent zero-shot reconstruction methods, and methods exploiting diffusion priors, with DM-ZS serving as the closest baseline to DIDM. The parameters of each competing method are set according to the configurations recommended in the corresponding publication to ensure a consistent comparison.

\noindent\textbf{Evaluation Metrics.}
For RR assessment, six full-reference metrics are used for quantitative evaluation: peak signal-to-noise ratio (PSNR) \cite{Wald2000Quality}, root mean square error (RMSE) \cite{Wald2000Quality}, spectral angle mapper (SAM) \cite{Kruse1993SAM}, erreur relative globale adimensionnelle de synth\`ese (ERGAS) \cite{Wald2000Quality}, structural similarity (SSIM) \cite{Wang2004SSIM}, and correlation coefficient (CC) \cite{Vivone2015Critical}. PSNR and RMSE assess pixel-wise fidelity, SAM measures spectral distortion, ERGAS reflects global relative error, and SSIM and CC characterize structural similarity and correlation. Higher PSNR, SSIM, and CC and lower RMSE, SAM, and ERGAS indicate better performance. For FR assessment, we report the spectral distortion index $D_{\lambda}$, spatial distortion index $D_S$, and hybrid quality with no reference (HQNR) \cite{vivone2022panchromatic,Arienzo2022FRQA}. Lower $D_{\lambda}$ and $D_S$ indicate smaller spectral and spatial distortions, whereas higher HQNR indicates better overall no-reference quality. Since FR1 has no reference HRHS, full-reference metrics are not computed. RR results are averaged over all crops of each dataset, while $D_{\lambda}$, $D_S$, and HQNR are reported for the single FR1 sample.

\noindent\textbf{Implementation Details.} DIDM is implemented in PyTorch, and all experiments are conducted on a single NVIDIA GeForce RTX 4090D GPU. The configurations are summarized as follows for reproducibility: 1) The remote-sensing pre-trained diffusion backbone is kept frozen except for the cross-attention layers introduced for prompt injection; the prompt modules and spatial--spectral reconstruction module are jointly optimized. 2) The number of reverse diffusion steps is set to $T=200$, and the NSSD reconstruction rank $r$ is set to 20, 40, 40, and 20 for Pavia, Chikusei, Houston, and FR1, respectively. 3) The spectral-token ratio $\alpha$ is fixed at the dataset level, with $\alpha=0.3$ for Pavia and $\alpha=0.5$ for Chikusei, Houston, and FR1. 4) The loss weights are set to $\lambda_{\mathrm{spe}}=1$, $\lambda_{\mathrm{spa}}=1$, and $\lambda_{\mathrm{str}}=50$, and the WPATV winsorization percentile $p$ is set to 0.995. All learnable parameters are optimized using Adam with a fixed learning rate of $1\times10^{-3}$. 5) DIDM follows instance-wise zero-shot optimization with a batch size of one, and each LRHS/PAN pair is optimized independently for 25,000 iterations. 6) FR1 follows the same dataset-level configuration strategy as the RR experiments, without additional FR-specific ablations or exhaustive hyperparameter search. Unless otherwise stated, these settings are used throughout the experiments.

\subsection{Comparison with State-of-the-Art Methods}

To provide a comprehensive evaluation, we compare DIDM with representative methods under two complementary protocols. The RR assessment quantifies reconstruction fidelity on Pavia, Chikusei, and Houston using reference HRHS images, whereas the FR assessment evaluates fusion behavior on real multiresolution FR1 observations using $D_{\lambda}$, $D_S$, HQNR, and qualitative comparison. Bicubic denotes the LRHS upsampled to the target spatial resolution by bicubic interpolation and is included as a non-fusion reference. These two protocols jointly characterize reference-based reconstruction accuracy and native-resolution fusion quality.

\subsubsection{Reduced-Resolution Assessment}

\begin{figure*}[!t]
    \centering
    \includegraphics[width=\textwidth]{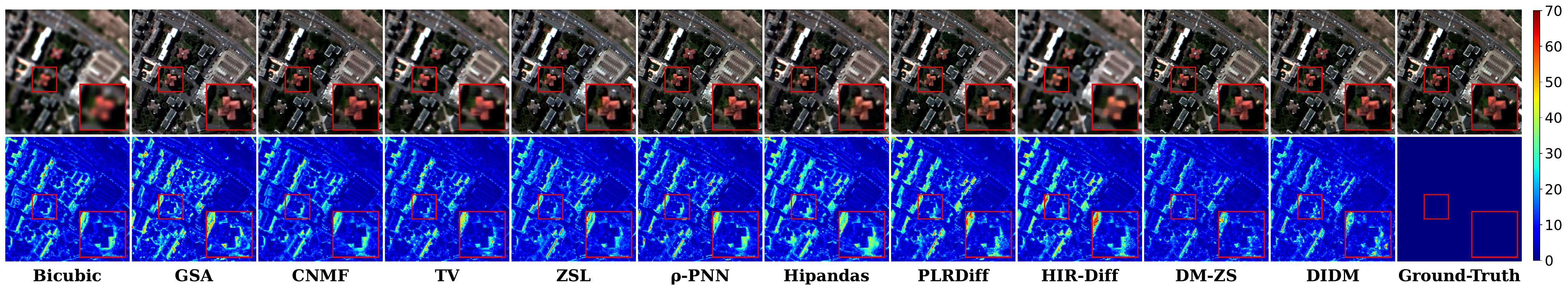}
    \caption{Visual comparison on the Pavia dataset. The pseudo-color images are generated using bands 50, 27, and 17 as the R, G, and B channels, respectively. The first row shows the reconstruction results, and the second row shows the corresponding error maps. Red rectangles indicate the selected regions of interest (ROIs) and their enlarged views.}
    \label{fig:pavia_visual}
\end{figure*}

\begin{figure*}[!t]
    \centering
    \includegraphics[width=\textwidth]{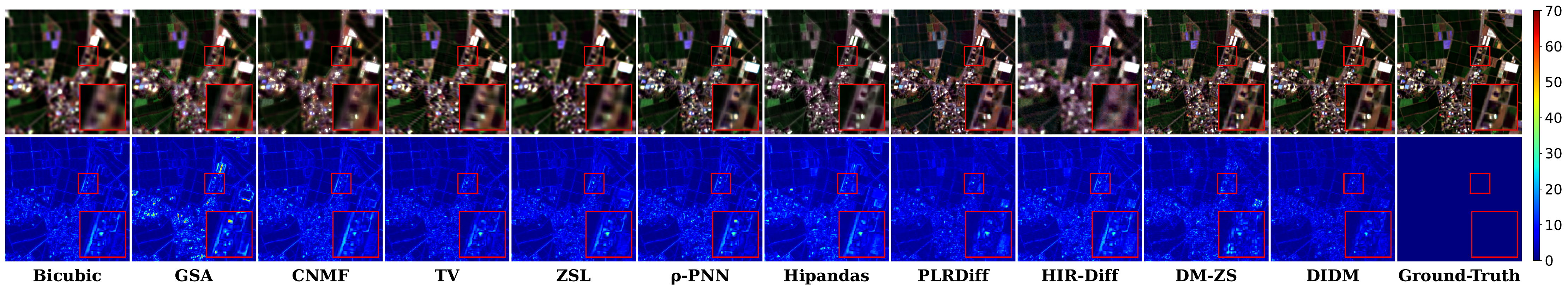}
    \caption{Visual comparison on the Chikusei dataset. The pseudo-color images are generated using bands 60, 40, and 20 as the R, G, and B channels, respectively. The first row shows the reconstruction results, and the second row shows the corresponding error maps. Red rectangles indicate the selected ROIs and their enlarged views.}
    \label{fig:chikusei_visual}
\end{figure*}

\begin{figure*}[!t]
    \centering
    \includegraphics[width=\textwidth]{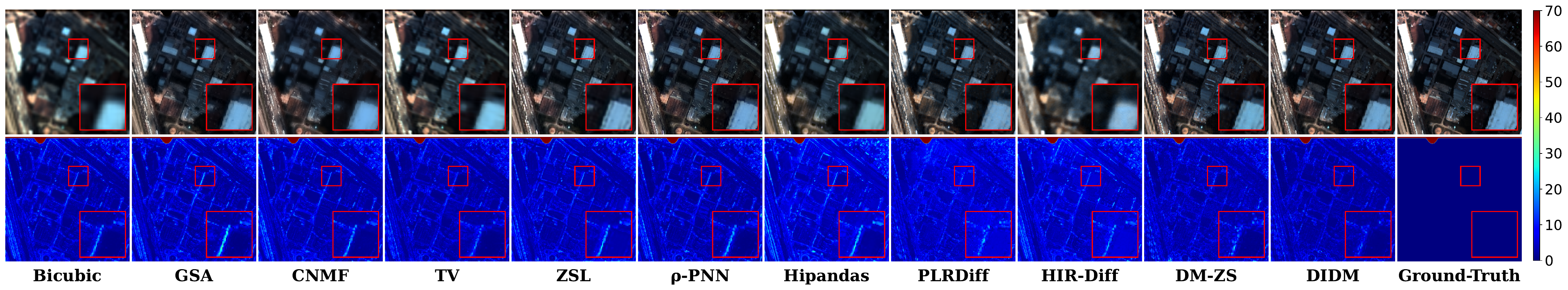}
    \caption{Visual comparison on the Houston dataset. The pseudo-color images are generated using bands 59, 31, and 16 as the R, G, and B channels, respectively. The first row shows the reconstruction results, and the second row shows the corresponding error maps. Red rectangles indicate the selected ROIs and their enlarged views.}
    \label{fig:houston_visual}
\end{figure*}

Tables~\ref{tab:pavia_comparison_revised}--\ref{tab:houston_comparison_revised} report the quantitative results on Pavia, Chikusei, and Houston, respectively. Figs.~\ref{fig:pavia_visual}--\ref{fig:houston_visual} present the corresponding pseudo-color reconstructions and error maps. Together, these results evaluate pixel-level reconstruction fidelity, spectral preservation, structural similarity, and spectral--spatial correlation under the RR reference-based protocol.

\begin{table}[!t]
\centering
\caption{Quantitative comparison on the Pavia dataset. The best and second-best results are highlighted in bold and underlined, respectively.}
\label{tab:pavia_comparison_revised}
\scriptsize
\setlength{\tabcolsep}{2.2pt}
\renewcommand{\arraystretch}{1.05}
\resizebox{\columnwidth}{!}{%
\begin{tabular}{ccccccc}
\hline
Method & PSNR $\uparrow$ & RMSE $\downarrow$ & SAM $\downarrow$ & ERGAS $\downarrow$ & SSIM $\uparrow$ & CC $\uparrow$ \\
\hline
Bicubic & 26.833 & 4.628 & 7.238 & 8.392 & 0.673 & 0.895 \\
GSA & 25.052 & 5.819 & 11.441 & 12.763 & 0.776 & 0.972 \\
CNMF & 25.369 & 5.671 & \underline{7.158} & 9.689 & 0.882 & 0.975 \\
TV & 22.172 & 8.010 & 8.228 & 14.646 & 0.739 & 0.949 \\
ZSL & 32.168 & 2.494 & 7.679 & 4.753 & 0.901 & 0.970 \\
$\rho$-PNN & 32.779 & 2.333 & 7.201 & 4.418 & 0.915 & 0.974 \\
Hipandas & 30.181 & 3.147 & 8.649 & 5.774 & 0.853 & 0.954 \\
PLRDiff & 33.527 & 2.133 & 8.389 & 4.316 & 0.916 & 0.978 \\
HIR-Diff & 27.509 & 4.289 & 7.878 & 7.878 & 0.699 & 0.908 \\
DM-ZS & \underline{33.766} & \underline{2.080} & 7.295 & \underline{4.115} & \underline{0.918} & \underline{0.979} \\
\textbf{DIDM} & \textbf{34.932} & \textbf{1.819} & \textbf{6.283} & \textbf{3.640} & \textbf{0.937} & \textbf{0.984} \\
\hline
\end{tabular}%
}
\end{table}

On Pavia, DIDM achieves the best result for every reported metric. Compared with DM-ZS, which remains the strongest competitor in overall reconstruction accuracy, DIDM improves PSNR from 33.766 to 34.932 dB and reduces RMSE from 2.080 to 1.819. Among the direct zero-shot reconstruction methods, $\rho$-PNN performs best overall with 32.779 dB PSNR and 7.201 SAM, whereas DIDM further improves these values to 34.932 dB and 6.283. Bicubic preserves a relatively low SAM but has substantially lower PSNR and SSIM, reflecting the absence of high-resolution spatial detail. The consistent gains across error-, spectral-, and structure-oriented metrics indicate that DIDM recovers urban structures without sacrificing spectral fidelity.

\begin{table}[!t]
\centering
\caption{Quantitative comparison on the Chikusei dataset. The best and second-best results are highlighted in bold and underlined, respectively.}
\label{tab:chikusei_comparison_revised}
\scriptsize
\setlength{\tabcolsep}{2.2pt}
\renewcommand{\arraystretch}{1.05}
\resizebox{\columnwidth}{!}{%
\begin{tabular}{ccccccc}
\hline
Method & PSNR $\uparrow$ & RMSE $\downarrow$ & SAM $\downarrow$ & ERGAS $\downarrow$ & SSIM $\uparrow$ & CC $\uparrow$ \\
\hline
Bicubic & 36.056 & 1.588 & 3.206 & 7.063 & 0.893 & 0.976 \\
GSA & 22.541 & 7.874 & 11.539 & 23.909 & 0.740 & \textbf{0.990} \\
CNMF & 18.707 & 12.130 & 3.677 & 13.438 & 0.795 & 0.981 \\
TV & 21.730 & 8.692 & 5.547 & 21.570 & 0.692 & 0.984 \\
ZSL & 36.684 & 1.477 & 3.409 & 6.421 & 0.904 & 0.979 \\
$\rho$-PNN & 38.191 & 1.245 & \underline{3.183} & 5.043 & 0.929 & 0.985 \\
Hipandas & 36.489 & 1.510 & 4.268 & 7.980 & 0.906 & 0.978 \\
PLRDiff & \underline{39.029} & \underline{1.127} & 3.453 & 4.753 & \underline{0.938} & \underline{0.988} \\
HIR-Diff & 36.277 & 1.548 & 4.031 & 7.685 & 0.886 & 0.977 \\
DM-ZS & 37.700 & 1.316 & 3.484 & \underline{4.616} & 0.921 & 0.984 \\
\textbf{DIDM} & \textbf{39.797} & \textbf{1.031} & \textbf{2.740} & \textbf{3.629} & \textbf{0.949} & \textbf{0.990} \\
\hline
\end{tabular}%
}
\end{table}

On Chikusei, DIDM obtains the best PSNR, RMSE, SAM, ERGAS, and SSIM, while tying for the highest CC. PLRDiff remains the strongest competitor in PSNR and RMSE, whereas $\rho$-PNN provides the second-best SAM of 3.183. DIDM improves PSNR over PLRDiff by 0.768 dB and reduces SAM from 3.183 to 2.740 relative to $\rho$-PNN. Bicubic preserves the coarse spectral content reasonably well, but its lower PSNR and SSIM indicate limited spatial-detail recovery, while Hipandas remains below $\rho$-PNN on the main spectral and structural measures. These results support a more balanced use of spectral and spatial observations in scenes with dense field patterns and fine linear structures.

\begin{table}[!t]
\centering
\caption{Quantitative comparison on the Houston dataset. The best and second-best results are highlighted in bold and underlined, respectively.}
\label{tab:houston_comparison_revised}
\scriptsize
\setlength{\tabcolsep}{2.2pt}
\renewcommand{\arraystretch}{1.05}
\resizebox{\columnwidth}{!}{%
\begin{tabular}{ccccccc}
\hline
Method & PSNR $\uparrow$ & RMSE $\downarrow$ & SAM $\downarrow$ & ERGAS $\downarrow$ & SSIM $\uparrow$ & CC $\uparrow$ \\
\hline
Bicubic & 31.788 & 2.679 & 6.140 & 6.512 & 0.789 & 0.916 \\
GSA & 30.745 & 3.028 & 9.130 & 10.188 & 0.848 & 0.975 \\
CNMF & 24.537 & 7.815 & 6.400 & 22.704 & 0.767 & 0.968 \\
TV & 20.267 & 10.327 & 6.597 & 30.731 & 0.653 & 0.945 \\
ZSL & 35.409 & 1.760 & 6.340 & 4.334 & 0.887 & 0.963 \\
$\rho$-PNN & 37.396 & 1.418 & 6.077 & 3.495 & 0.917 & 0.977 \\
Hipandas & 34.173 & 2.046 & 7.425 & 5.338 & 0.862 & 0.951 \\
PLRDiff & 38.106 & 1.294 & 6.265 & 3.445 & 0.931 & 0.980 \\
HIR-Diff & 32.045 & 2.587 & 6.891 & 6.371 & 0.788 & 0.920 \\
DM-ZS & \underline{38.435} & \underline{1.247} & \underline{5.539} & \underline{3.178} & \underline{0.932} & \underline{0.982} \\
\textbf{DIDM} & \textbf{39.172} & \textbf{1.152} & \textbf{5.073} & \textbf{2.896} & \textbf{0.939} & \textbf{0.985} \\
\hline
\end{tabular}%
}
\end{table}

On Houston, DIDM again ranks first across all six metrics. DM-ZS remains the strongest overall competitor, with DIDM increasing PSNR from 38.435 to 39.172 dB, reducing SAM from 5.539 to 5.073, and lowering ERGAS from 3.178 to 2.896. $\rho$-PNN is the strongest direct zero-shot reconstruction baseline, reaching 37.396 dB PSNR and 6.077 SAM, while Hipandas and Bicubic show lower overall reconstruction accuracy. Given the dense man-made structures and pronounced spectral variability of Houston, the consistent gains support the effectiveness of coupling internal dual-modality prompting with PAN-guided structural regularization.

Overall, DIDM achieves the best values for all reported RR metrics across the three datasets, even after including Bicubic and the recent $\rho$-PNN and Hipandas baselines. The improvements extend beyond PSNR to SAM, ERGAS, SSIM, and CC, demonstrating a consistent balance between spatial-detail recovery and spectral preservation under the controlled RR protocol.

The visual comparisons are consistent with the quantitative results. Bicubic largely preserves the coarse spectral appearance of LRHS but lacks high-resolution structures, leading to blurred boundaries and stronger residuals. The classical methods recover part of the missing spatial information, while the direct zero-shot reconstruction methods, including ZSL, $\rho$-PNN, and Hipandas, recover richer structures with different levels of sharpness. The diffusion-prior methods enhance high-frequency details, although some results exhibit residual blur, pseudo-textures, or stronger error responses. These differences indicate that apparent sharpness alone does not necessarily yield a spatially and spectrally balanced reconstruction.

On Pavia, DIDM better delineates roof contours, road intersections, and shadow boundaries. Bicubic substantially smooths these structures, while the classical methods recover their coarse geometry but leave visible boundary residuals. $\rho$-PNN and Hipandas recover more spatial detail than the smoother zero-shot results, yet residual responses remain around several high-contrast structures. The diffusion-prior methods provide sharper urban details, whereas DIDM produces local transitions closer to the ground truth with relatively compact residual responses in the selected ROI.

On Chikusei, Bicubic noticeably weakens the fine field patterns and narrow linear structures. GSA, CNMF, TV, and ZSL recover the large-scale field layout but leave some boundaries blurred or attenuated. $\rho$-PNN and Hipandas improve local structural recovery, although residual responses remain along several fine boundaries. The diffusion-prior methods preserve more high-frequency content, while DIDM maintains the field grids, narrow linear features, and agricultural textures with more regular spatial transitions and relatively weak residual responses in the enlarged ROI.

On Houston, Bicubic produces a smooth reconstruction with reduced fine structural contrast, whereas the classical and direct zero-shot reconstruction methods progressively recover building contours and linear structures. $\rho$-PNN and Hipandas retain more local detail than the interpolation reference, while the diffusion-prior methods further improve structural sharpness with different levels of texture regularity. DIDM preserves clearer transitions among roofs, roads, and shadowed regions, with relatively compact residual responses around the selected ROI. This behavior is consistent with the complementary roles of LRHS-derived spectral prompting, PAN-derived spatial prompting, and PAN-guided WPATV.

\subsubsection{Full-Resolution Evaluation on FR1}

\begin{figure*}[!t]
    \centering
    \includegraphics[width=\textwidth]{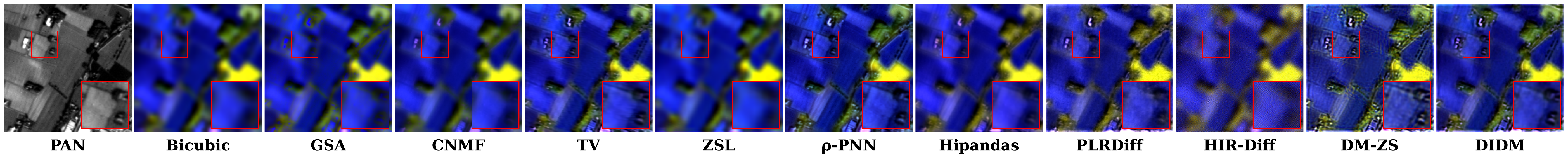}
    \caption{Visual comparison on the released FR1 full-resolution sample. From left to right: PAN, Bicubic, GSA, CNMF, TV, ZSL, $\rho$-PNN, Hipandas, PLRDiff, HIR-Diff, DM-ZS, and DIDM. Bicubic denotes the LRHS upsampled to the PAN scale by bicubic interpolation. The pseudo-color images are generated using bands 20, 30, and 40 as the R, G, and B channels, respectively. Red rectangles indicate the selected ROI, whose enlarged view is shown at the bottom right of each image.}
    \label{fig:fr1_visual}
\end{figure*}

The FR1 experiment complements the RR assessment by evaluating the methods on real PAN and hyperspectral observations at their native resolutions. Since FR1 does not provide a reference HRHS image, Table~\ref{tab:fr1_comparison} reports $D_{\lambda}$, $D_S$, and HQNR, while Fig.~\ref{fig:fr1_visual} presents the corresponding visual comparison. The analysis focuses on spectral and spatial distortion, overall no-reference fusion quality, PAN-aligned structural detail, and artifact suppression.

\begin{table*}[!t]
\centering
\caption{Full-resolution quantitative comparison on the released FR1 sample. The best and second-best results are highlighted in bold and underlined, respectively. Rankings are determined from the original unrounded values.}
\label{tab:fr1_comparison}
\scriptsize
\setlength{\tabcolsep}{4.2pt}
\renewcommand{\arraystretch}{1.08}
\begin{tabular}{cccccccccccc}
\toprule
Metric & Bicubic & GSA & CNMF & TV & ZSL & $\rho$-PNN & Hipandas & PLRDiff & HIR-Diff & DM-ZS & DIDM \\
\midrule
$D_{\lambda}\downarrow$
& \textbf{0.021} & 0.042 & 0.063 & \underline{0.022} & 0.110
& 0.048 & 0.092 & 0.060 & 0.099 & 0.044 & 0.040 \\

$D_S\downarrow$
& 0.225 & 0.143 & 0.100 & 0.150 & \textbf{0.056}
& 0.148 & \underline{0.086} & 0.121 & 0.114 & 0.118 & 0.108 \\

HQNR $\uparrow$
& 0.759 & 0.821 & \underline{0.843} & 0.832 & 0.840
& 0.812 & 0.830 & 0.826 & 0.799 & 0.843 & \textbf{0.857} \\
\bottomrule
\end{tabular}
\end{table*}

As shown in Table~\ref{tab:fr1_comparison}, DIDM achieves the highest HQNR of 0.857 on FR1, demonstrating the best overall no-reference fusion quality among the evaluated methods. Bicubic and TV yield the two lowest $D_{\lambda}$ values of 0.021 and 0.022, respectively, but their spatial distortions are considerably larger; in particular, Bicubic exhibits the highest $D_S$ of 0.225 and the lowest HQNR of 0.759, reflecting its inability to recover the missing high-resolution spatial information. Conversely, ZSL achieves the lowest $D_S$ of 0.056 but has the largest $D_{\lambda}$ of 0.110, while Hipandas provides the second-lowest $D_S$ of 0.086. DIDM maintains relatively low distortions in both dimensions, with $D_{\lambda}=0.040$ and $D_S=0.108$, and outperforms the other diffusion-prior methods on both distortion indices. Its highest HQNR therefore reflects a more favorable balance between spectral preservation and spatial-detail enhancement rather than an advantage in only one distortion component.

Taking PAN as the reference for spatial structure and Bicubic as the reference for the coarse spectral appearance of LRHS, Fig.~\ref{fig:fr1_visual} reveals distinct fusion behaviors among the compared methods. Bicubic preserves the overall pseudo-color distribution but lacks fine spatial details. Among the classical methods, GSA and CNMF remain comparatively smooth around field boundaries, whereas TV introduces stronger spatial enhancement with locally over-emphasized textures. ZSL also produces a relatively smooth reconstruction. The recent direct zero-shot reconstruction methods improve spatial recovery: $\rho$-PNN delineates field boundaries and narrow structures more clearly, while Hipandas provides smoother local transitions. The diffusion-prior methods recover richer high-frequency patterns, although PLRDiff and DM-ZS exhibit visible grain-like or grid-like fluctuations in some homogeneous regions, and HIR-Diff shows prominent high-frequency artifacts and local chromatic variation in the enlarged ROI.

In contrast, DIDM produces a cleaner and more structurally coherent result on FR1. Field boundaries and narrow linear structures are clearly delineated, while homogeneous agricultural regions contain fewer spurious high-frequency fluctuations. The enlarged ROI further shows sharper yet more regular local textures than the smoother reconstruction methods, without the pronounced grain-like responses observed in highly sharpened results. Meanwhile, the overall pseudo-color appearance remains consistent with the LRHS observation. These visual characteristics are consistent with the joint effect of dual-modality prompt injection and PAN-guided WPATV. Together with the highest HQNR and low spectral distortion, the results support a favorable balance between spatial-detail enhancement and spectral preservation on FR1.

\subsection{Ablation Studies and Analysis}

Having established the comparative performance of DIDM, we next investigate the sources of its gains and the practical implications of its instance-wise zero-shot formulation. We first isolate the contribution of each core component and examine whether prompt injection benefits from observation content rather than additional optimized parameters. We then analyze the prompt balance, injection positions, modality-specific responses, and PAN-guided WPATV. Finally, we characterize the computational behavior from two complementary perspectives: the per-pair optimization budget and the reverse-diffusion trajectory. To match the purpose of each experiment, discrete component and injection-position ablations report all six RR metrics, whereas continuous parameter sweeps use PSNR and SAM as compact representatives of reconstruction fidelity and spectral distortion, respectively. In the efficiency analysis, the iteration--cost experiment visualizes PSNR against runtime and reports SAM in the text, while the reverse-step analysis reports both PSNR and SAM to characterize sampling saturation.

\subsubsection{Component Analysis and Parameter Budget}

We conduct the component analysis on Houston because its complex urban structures and pronounced spectral variability provide a challenging setting for jointly evaluating spatial-detail reconstruction and spectral preservation. Table~\ref{tab:component_ablation} summarizes the variants. The \emph{Baseline} removes both the feature-level prompt injection and the PAN-guided WPATV while retaining the remaining reconstruction framework and observation-fidelity terms, whereas \emph{w/o Prompt Inj.} removes the prompt injection but preserves the WPATV and the fidelity objective. The \emph{w/o Spe. Token} and \emph{w/o Spa. Token} variants suppress the LRHS-derived spectral token and the PAN-derived spatial token, respectively, and \emph{w/o WPATV} removes only the structural regularizer. To control for the additional capacity introduced by prompt injection, \emph{w/ Content-Free Prompt} retains the complete prompt-injection architecture and optimization objective but replaces the prompt-encoder inputs with constants, setting every LRHS voxel and PAN pixel to 0.5.

\begin{table}[!t]
\centering
\caption{Component ablation and parameter-capacity control on the Houston dataset. The best results are in boldface.}
\label{tab:component_ablation}
\scriptsize
\setlength{\tabcolsep}{2.0pt}
\renewcommand{\arraystretch}{1.05}
\resizebox{\columnwidth}{!}{%
\begin{tabular}{ccccccc}
\hline
Variant & PSNR $\uparrow$ & RMSE $\downarrow$ & SAM $\downarrow$ & ERGAS $\downarrow$ & SSIM $\uparrow$ & CC $\uparrow$ \\
\hline
Baseline & 38.398 & 1.264 & 5.757 & 3.196 & 0.927 & 0.981 \\
w/o Prompt Inj. & 38.915 & 1.172 & 5.371 & 2.974 & 0.937 & 0.984 \\
w/ Content-Free Prompt & 38.948 & 1.185 & 5.170 & 2.973 & 0.937 & 0.984 \\
w/o Spe. Token & 39.068 & 1.157 & 5.185 & 2.929 & 0.937 & 0.984 \\
w/o Spa. Token & 39.008 & 1.157 & 5.153 & 2.952 & 0.937 & 0.984 \\
w/o WPATV & 38.731 & 1.159 & 5.512 & 3.067 & 0.933 & 0.983 \\
\textbf{DIDM} & \textbf{39.172} & \textbf{1.152} & \textbf{5.073} & \textbf{2.896} & \textbf{0.939} & \textbf{0.985} \\
\hline
\end{tabular}%
}
\end{table}

The complete DIDM configuration achieves the best result for every metric, confirming the complementary contributions of its components. Relative to the \emph{Baseline}, DIDM improves PSNR by 0.774 dB and reduces SAM from 5.757 to 5.073. Removing prompt injection lowers PSNR from 39.172 to 38.915 dB and increases SAM from 5.073 to 5.371, showing that injecting LRHS/PAN observations into diffusion feature evolution provides useful conditioning beyond the target-level fidelity objective. Removing either modality-specific token also degrades performance: PSNR decreases to 39.068 dB without the spectral token and 39.008 dB without the spatial token, supporting the complementarity of LRHS-derived spectral cues and PAN-derived spatial structures. Removing WPATV further reduces PSNR to 38.731 dB and increases SAM and ERGAS to 5.512 and 3.067, respectively, demonstrating its role in suppressing structure-related distortions while preserving observation fidelity.

The content-free control separates observation content from parameter capacity. DM-ZS contains 393.069 million total parameters, of which 1.396 million are optimized, whereas DIDM contains 449.239 million total parameters and optimizes 58.191 million parameters; the increase mainly comes from the active cross-attention layers and two image-projection modules used for prompt injection. Despite retaining the same prompt-injection architecture and optimized parameter budget as DIDM, the content-free variant remains inferior on all six metrics, obtaining 38.948 dB PSNR and 5.170 SAM. Relative to \emph{w/o Prompt Inj.}, its changes are small and metric-dependent: PSNR increases by only 0.033 dB, RMSE increases from 1.172 to 1.185, and SSIM and CC remain unchanged. By contrast, supplying the prompt pathway with the actual LRHS and PAN observations yields consistent improvements across all metrics, indicating that the gains of DIDM arise primarily from observation-aware dual-modality prompting rather than from the enlarged optimized parameter budget alone.

\subsubsection{Prompt Design Analysis}

After establishing the effectiveness of the observation-aware prompt injection, we analyze the prompt design from three perspectives: the balance between LRHS-derived spectral and PAN-derived spatial prompts, the feature levels at which they are injected, and their modality-specific responses within the diffusion backbone. These experiments examine whether the two prompts are complementary, whether conditioning should span multiple feature scales, and whether the branches exhibit distinct internal behaviors.

\begin{figure*}[!t]
    \centering
    \includegraphics[width=0.95\textwidth]{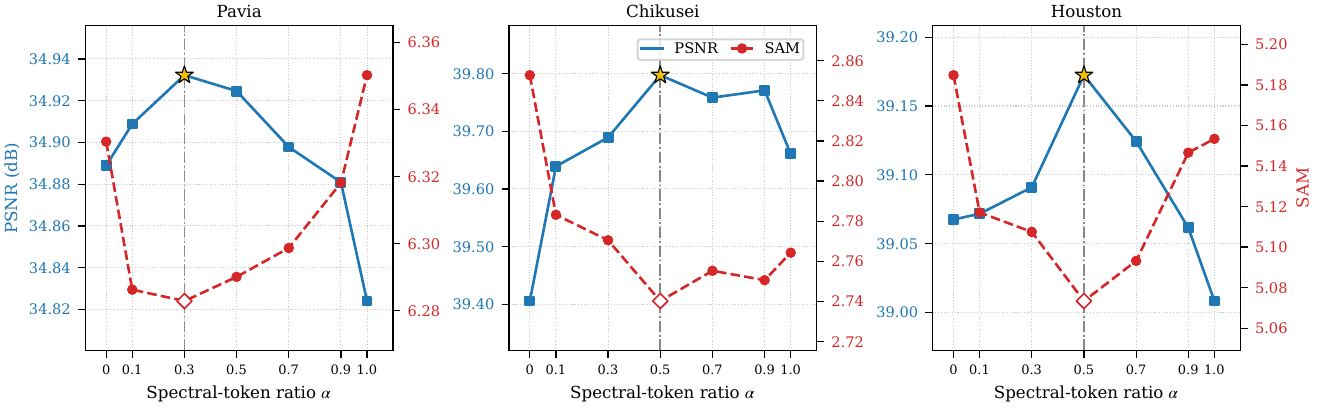}
    \caption{Effect of the spectral-token ratio $\alpha$ on Pavia, Chikusei, and Houston under the RR reference-based setting. The two endpoints, $\alpha=0$ and $\alpha=1$, correspond to spatial-only and spectral-only prompting, respectively.}
    \label{fig:injection_alpha_all}
\end{figure*}

\noindent\textbf{Prompt Balance.} The spectral-token ratio $\alpha$ controls the relative contribution of the two modality-specific prompts: $\alpha=0$ retains only the PAN-derived spatial prompt, $\alpha=1$ retains only the LRHS-derived spectral prompt, and intermediate values combine both modalities. To compactly characterize the reconstruction and spectral trends across three datasets, Fig.~\ref{fig:injection_alpha_all} reports PSNR and SAM as representative RR metrics. The results show that joint prompting is consistently more effective than either single-modality alternative, although the preferred balance varies across datasets. (a) On Houston, spatial-only and spectral-only prompting yield PSNR/SAM values of 39.068/5.185 and 39.008/5.153, respectively, both below 39.172/5.073 at $\alpha=0.5$. (b) A moderate spectral--spatial balance is generally favorable: Chikusei achieves its best PSNR/SAM at $\alpha=0.5$ (39.797/2.740), whereas Pavia favors the slightly more spatially weighted setting $\alpha=0.3$ (34.932/6.283). (c) The improvement of intermediate ratios over the two endpoints confirms that LRHS-derived spectral evidence and PAN-derived spatial structures provide complementary conditioning; accordingly, the selected dataset-level value is fixed for all samples in the main experiments.

\noindent\textbf{Injection Positions.} We examine whether prompt conditioning should be restricted to a specific feature level by independently enabling or disabling prompt injection in the downsampling path (\emph{Down}), middle block (\emph{Mid}), and upsampling path (\emph{Up}), which represent different feature resolutions and abstraction levels in the U-Net-like diffusion backbone. The Houston results in Table~\ref{tab:injection_position} show that prompt injection is beneficial at every stage and that distributing it across the full diffusion hierarchy yields the strongest overall performance. (a) Each single-stage configuration improves upon the variant without feature-level prompting: the three single-stage variants achieve PSNR values of 39.091--39.097 dB and reduce SAM from 5.371 to 5.092--5.133, indicating that the LRHS/PAN observations provide useful conditioning at different feature resolutions. (b) Multi-stage injection is generally more effective than single-stage injection, as demonstrated by the \emph{Down+Mid} and \emph{Mid+Up} configurations, which reduce SAM to 5.075 and 5.078, respectively. (c) Jointly activating \emph{Down}, \emph{Mid}, and \emph{Up} achieves the best result, improving PSNR from 38.915 to 39.172 dB and reducing SAM from 5.371 to 5.073 relative to disabling all injection stages, which indicates that propagating the observation prompts through the encoding, bottleneck, and reconstruction paths better exploits the multiscale feature hierarchy than restricting them to a single level.

\begin{table}[!t]
\centering
\caption{Effect of prompt-injection positions on the Houston dataset. $\checkmark$ and $\times$ indicate enabled and disabled prompt injection, respectively. The best results are in boldface.}
\label{tab:injection_position}
\scriptsize
\setlength{\tabcolsep}{2.0pt}
\renewcommand{\arraystretch}{1.05}
\resizebox{\columnwidth}{!}{%
\begin{tabular}{ccccccccc}
\hline
Down & Mid & Up & PSNR $\uparrow$ & RMSE $\downarrow$ & SAM $\downarrow$ & ERGAS $\downarrow$ & SSIM $\uparrow$ & CC $\uparrow$ \\
\hline
$\times$ & $\times$ & $\times$ & 38.915 & 1.172 & 5.371 & 2.974 & 0.937 & 0.984 \\
$\checkmark$ & $\times$ & $\times$ & 39.097 & 1.162 & 5.092 & 2.920 & 0.939 & 0.984 \\
$\times$ & $\checkmark$ & $\times$ & 39.091 & 1.162 & 5.133 & 2.921 & 0.939 & 0.984 \\
$\times$ & $\times$ & $\checkmark$ & 39.096 & 1.164 & 5.107 & 2.920 & 0.938 & 0.984 \\
$\checkmark$ & $\checkmark$ & $\times$ & 39.116 & 1.160 & 5.075 & 2.912 & 0.939 & 0.984 \\
$\checkmark$ & $\times$ & $\checkmark$ & 39.120 & 1.159 & 5.118 & 2.912 & 0.939 & 0.984 \\
$\times$ & $\checkmark$ & $\checkmark$ & 39.128 & 1.159 & 5.078 & 2.914 & 0.939 & 0.984 \\
\textbf{$\checkmark$} & \textbf{$\checkmark$} & \textbf{$\checkmark$} &
\textbf{39.172} & \textbf{1.152} & \textbf{5.073} & \textbf{2.896} &
\textbf{0.939} & \textbf{0.985} \\
\hline
\end{tabular}%
}
\end{table}

\begin{figure}[!t]
    \centering
    \includegraphics[width=\columnwidth]{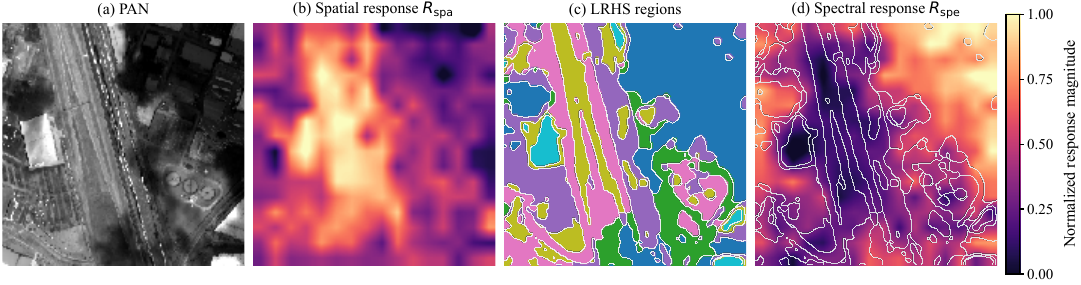}
    \caption{Visualization of modality-specific prompt responses on a Houston sample. (a) PAN observation. (b) Spatial-prompt response $R_{\mathrm{spa}}$. (c) LRHS-derived spectral regions obtained by clustering the LRHS spectral vectors. (d) Spectral-prompt response $R_{\mathrm{spe}}$, where the overlaid contours indicate the boundaries of the same LRHS-derived spectral partition shown in (c). The two response maps are independently normalized to $[0,1]$ for visualization; brighter colors indicate stronger relative responses within each map.}
    \label{fig:prompt_response}
\end{figure}

\noindent\textbf{Prompt-Response Behavior.} Fig.~\ref{fig:prompt_response} visualizes prompt-induced response magnitudes, rather than raw attention weights, extracted from the mid-block cross-attention outputs on a Houston sample. LRHS-derived spectral regions are obtained by clustering the spatially aligned LRHS spectral vectors and provide a reference for interpreting region-wise response organization. (a) $R_{\mathrm{spa}}$ is strongly activated along the dominant linear structure and around several high-contrast objects in the PAN image, whereas $R_{\mathrm{spe}}$ exhibits broader region-wise variation and weaker edge-following behavior. (b) This distinction is quantified by $\rho_{\nabla}$, the Spearman correlation between a response map and the PAN-gradient magnitude, and $E_{\mathrm{edge}}$, the ratio between the mean response over the highest 10\% PAN-gradient pixels and that over the remaining pixels. For the displayed sample, $R_{\mathrm{spa}}$ obtains $\rho_{\nabla}=0.326$ and $E_{\mathrm{edge}}=1.122$, compared with $-0.323$ and 0.830 for $R_{\mathrm{spe}}$, supporting the structure-oriented role of the PAN-derived prompt. (c) Region-wise dependence is further measured by $\eta_{\mathrm{reg}}^{2}$, the fraction of response variance explained by the LRHS-derived spectral partition. $R_{\mathrm{spe}}$ achieves $\eta_{\mathrm{reg,spe}}^{2}=0.624$, exceeding $\eta_{\mathrm{reg,spa}}^{2}=0.458$; together with its smoother spatial distribution, this result indicates stronger organization by the LRHS spectral grouping. These quantities serve only as descriptive response diagnostics rather than fusion-quality metrics, and the observed differences support the complementary roles of the two prompt branches.

\subsubsection{PAN-Guided Structural Regularization Analysis}

Having verified the contribution of WPATV in the component ablation, we further analyze the structural regularizer from three complementary perspectives: the effect of its weight, its joint interaction with the prompt balance, and its edge-preserving behavior. The analysis follows a progressive logic from quantitative benefit to parameter robustness and, finally, to mechanism-level visual evidence.

\begin{figure*}[!t]
    \centering
    \includegraphics[width=\textwidth]{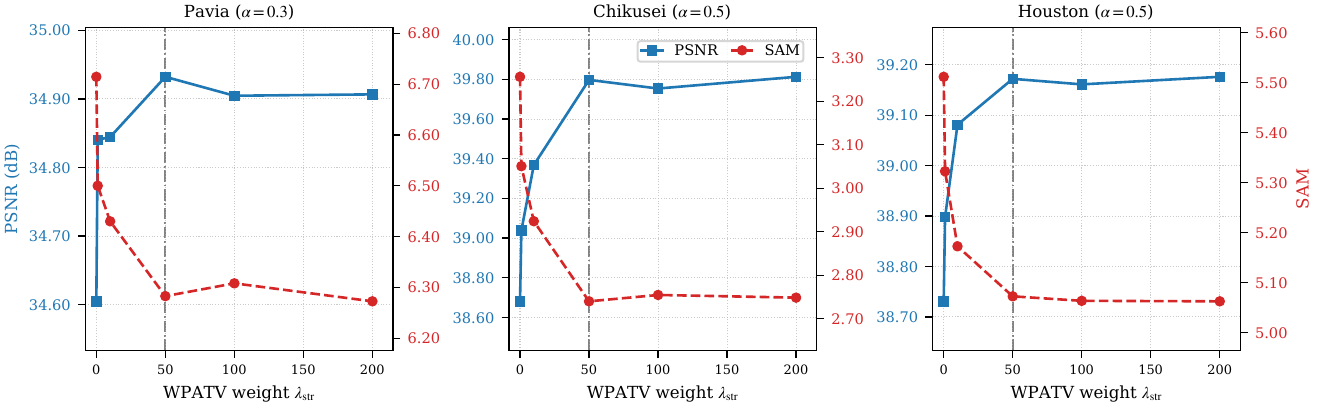}
    \caption{Effect of the PAN-guided WPATV weight $\lambda_{\mathrm{str}}$ on the Pavia, Chikusei, and Houston datasets. The spectral-token ratio $\alpha$ is fixed to $0.3$ for Pavia, and $0.5$ for both Chikusei and Houston.}
    \label{fig:lambda_wpatv_curve}
\end{figure*}

\noindent\textbf{Effect of the Structural Weight.} We examine the influence of the WPATV weight $\lambda_{\mathrm{str}}\in\{0,1,10,50,100,200\}$ on Pavia, Chikusei, and Houston while fixing $\alpha$ to the dataset-level settings of 0.3, 0.5, and 0.5, respectively; $\lambda_{\mathrm{str}}=0$ removes WPATV and, on Houston, corresponds to the \emph{w/o WPATV} configuration in Table~\ref{tab:component_ablation}. Because this experiment targets the spatial--spectral trend of a continuous hyperparameter sweep, Fig.~\ref{fig:lambda_wpatv_curve} reports PSNR and SAM as representative reconstruction and spectral metrics. A consistent improvement followed by saturation is observed across all three datasets. (a) Increasing $\lambda_{\mathrm{str}}$ from 0 to 50 raises PSNR from 34.605 to 34.932 dB, 38.682 to 39.797 dB, and 38.731 to 39.172 dB on Pavia, Chikusei, and Houston, respectively, while reducing SAM from 6.715 to 6.283, 3.256 to 2.740, and 5.512 to 5.073. (b) Beyond $\lambda_{\mathrm{str}}=50$, increasing the weight to 100 or 200 changes PSNR by at most 0.044 dB and SAM by at most 0.026 relative to the corresponding value at 50, indicating that the benefit largely saturates once an effective structural constraint is established. (c) We therefore adopt $\lambda_{\mathrm{str}}=50$ as a common setting because it reaches the onset of this stable regime and provides a favorable spatial--spectral balance without relying on stronger regularization. The cross-dataset trend supports WPATV as a robust structural complement to the observation-fidelity terms rather than a parameter requiring precise calibration.

\begin{figure*}[!t]
    \centering
    \includegraphics[width=\textwidth]{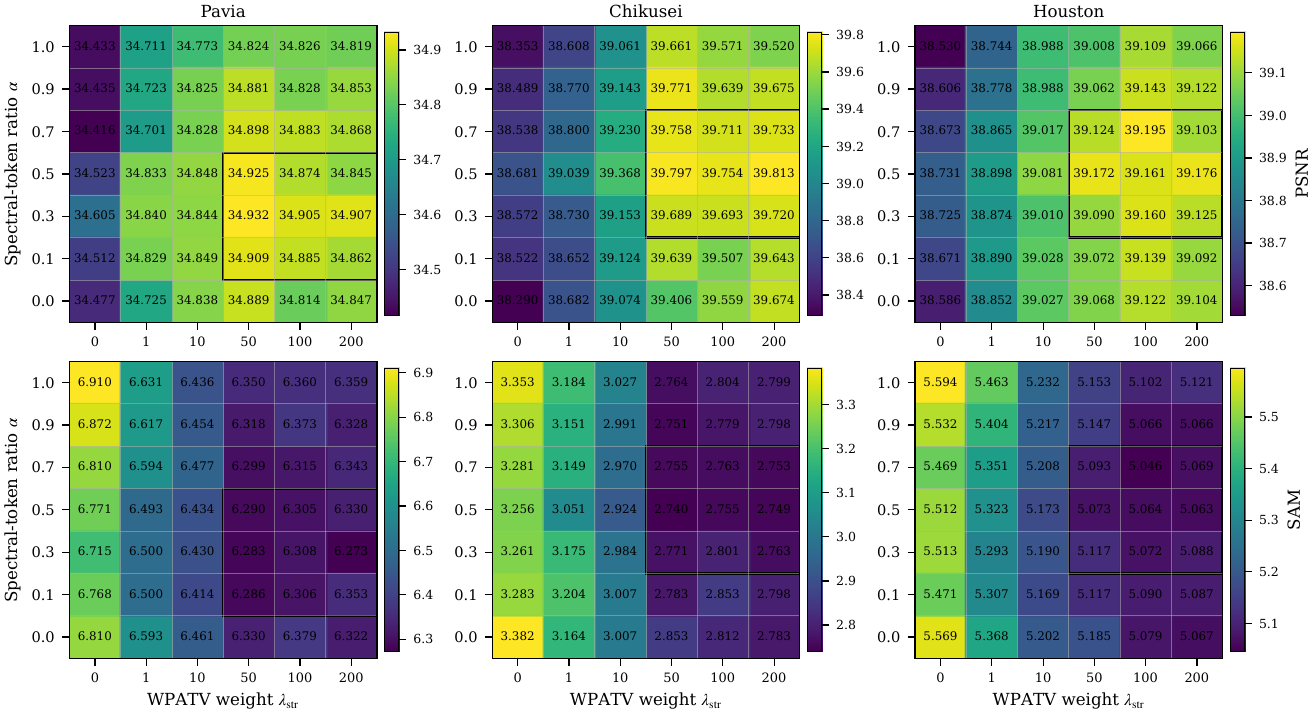}
    \caption{Joint sensitivity of the spectral-token ratio $\alpha$ and the WPATV weight $\lambda_{\mathrm{str}}$ across the Pavia, Chikusei, and Houston datasets in terms of PSNR (top row) and SAM (bottom row). The black rectangles mark stable high-performance regions for each specific dataset.}
    \label{fig:alpha_lambda_heatmap}
\end{figure*}

\noindent\textbf{Joint Sensitivity to Prompt Balance and Structural Regularization.} Because the spectral-token ratio $\alpha$ controls prompt composition whereas $\lambda_{\mathrm{str}}$ controls structural regularization, we jointly vary $\alpha\in\{0,0.1,0.3,0.5,0.7,0.9,1\}$ and $\lambda_{\mathrm{str}}\in\{0,1,10,50,100,200\}$. To keep the two-dimensional sensitivity analysis interpretable while covering both reconstruction fidelity and spectral distortion, Fig.~\ref{fig:alpha_lambda_heatmap} reports PSNR and SAM on all three datasets. The heatmaps show broad dataset-dependent high-performance regions rather than sharply isolated optima. (a) On Pavia, the black-boxed region spanning $\alpha=0.1$--0.5 and $\lambda_{\mathrm{str}}=50$--200 maintains PSNR within 34.845--34.932 dB and SAM within 6.273--6.353. (b) On Chikusei and Houston, the corresponding region of $\alpha=0.3$--0.7 and $\lambda_{\mathrm{str}}=50$--200 yields PSNR/SAM ranges of 39.689--39.813/2.740--2.801 and 39.090--39.195/5.046--5.117, respectively. (c) Although the exact optimum shifts among neighboring parameter combinations, the adopted settings $(\alpha,\lambda_{\mathrm{str}})=(0.3,50)$ for Pavia and $(0.5,50)$ for Chikusei and Houston lie within these stable regions and remain close to the best PSNR--SAM combinations. These results indicate that the two parameters play complementary roles without introducing a narrow operating range, supporting dataset-level selection of $\alpha$ together with the common choice $\lambda_{\mathrm{str}}=50$.

\begin{figure}[!t]
    \centering
    \includegraphics[width=\columnwidth]{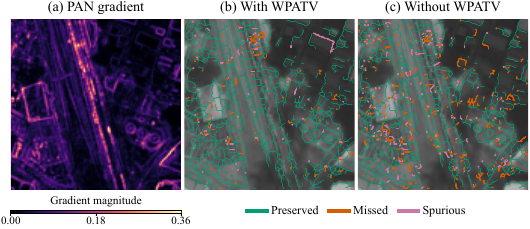}
    \caption{Visualization of PAN-aligned edge agreement on a Houston sample. (a) PAN-gradient magnitude. Edge-agreement maps obtained (b) with and (c) without WPATV. Green and vermillion curves denote preserved and missed PAN-derived reference edges, respectively, whereas magenta curves indicate spurious pseudo-PAN edges.}
    \label{fig:wpatv_edge}
\end{figure}

\noindent\textbf{Edge-Preservation Behavior.} To directly examine the structural effect of WPATV, Fig.~\ref{fig:wpatv_edge} compares PAN-derived reference edges with edges extracted from the pseudo-PAN projections of the fused results reconstructed with and without WPATV. The pseudo-PAN projections are generated using the same spectral response as the PAN observation, subjected to common radiometric normalization, and evaluated under identical edge-extraction and matching settings; green and vermillion curves denote preserved and missed PAN-derived reference edges, respectively, whereas magenta curves indicate spurious pseudo-PAN edges. The edge-agreement maps provide direct evidence of the structural benefit of WPATV. (a) Edge recall measures the fraction of PAN-reference edges retained by the reconstruction, edge precision measures the fraction of reconstructed edges aligned with the PAN reference, and edge F1 is their harmonic mean; these quantities are used only as structural diagnostics. (b) For the displayed sample, WPATV increases recall from 0.865 to 0.955, precision from 0.920 to 0.962, and F1 from 0.892 to 0.959, while reducing the numbers of missed and spurious edges from 903 to 298 and from 499 to 253, corresponding to reductions of 67.0\% and 49.3\%, respectively. (c) Consistently, the WPATV result contains fewer vermillion and magenta responses and more continuous green structures, supporting the intended weighting mechanism in which smoothing is relaxed around strong PAN transitions and strengthened in relatively homogeneous regions to preserve PAN-aligned discontinuities while suppressing structure-related artifacts.

\subsubsection{Optimization and Efficiency Analysis}

DIDM incurs per-pair computation through instance-wise zero-shot optimization and reverse-diffusion sampling. We therefore first examine the optimization budget that governs the overall adaptation cost and then analyze the reverse-diffusion trajectory that controls sampling refinement. This separation characterizes the diminishing returns associated with the two principal computational budgets of DIDM.

\begin{figure}[!t]
    \centering
    \includegraphics[width=\columnwidth]{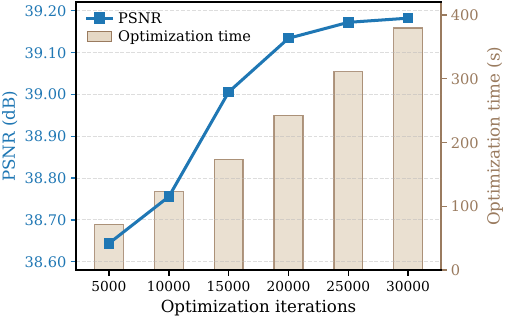}
    \caption{Performance-cost trade-off with respect to the number of per-pair optimization iterations on Houston. The line and bars report the mean PSNR and mean optimization time per sample, respectively.}
    \label{fig:iteration_cost}
\end{figure}

\noindent\textbf{Iteration--Cost Trade-off.} As an instance-wise zero-shot method, DIDM adapts its learnable modules independently to each LRHS/PAN pair. Because optimization-based and feed-forward methods follow different computational paradigms, we characterize the internal performance--cost behavior of DIDM rather than directly comparing runtime across method classes. Since this experiment focuses on the quality--cost relationship of per-pair adaptation, Fig.~\ref{fig:iteration_cost} visualizes PSNR as a representative reconstruction-quality metric together with optimization time, while SAM is reported in the text as a complementary spectral indicator. Over 5,000--30,000 iterations, reconstruction quality improves monotonically but gradually saturates. (a) Increasing the iteration count from 5,000 to 25,000 raises mean PSNR from 38.644 to 39.172 dB and reduces SAM from 5.413 to 5.073. (b) Beyond the adopted setting, increasing the budget from 25,000 to 30,000 iterations improves PSNR by only 0.010 dB and reduces SAM by 0.012, while the mean optimization time reaches 379.848 s, confirming that the performance gain has largely saturated. (c) We therefore adopt 25,000 iterations in the main comparisons as a practical balance between reconstruction quality and computational cost. Under this setting, the complete per-pair procedure takes approximately 331.1 s and reaches a peak GPU memory consumption of about 7.3 GB on a single NVIDIA GeForce RTX 4090D GPU.

\begin{figure*}[!t]
    \centering
    \includegraphics[width=0.95\textwidth]{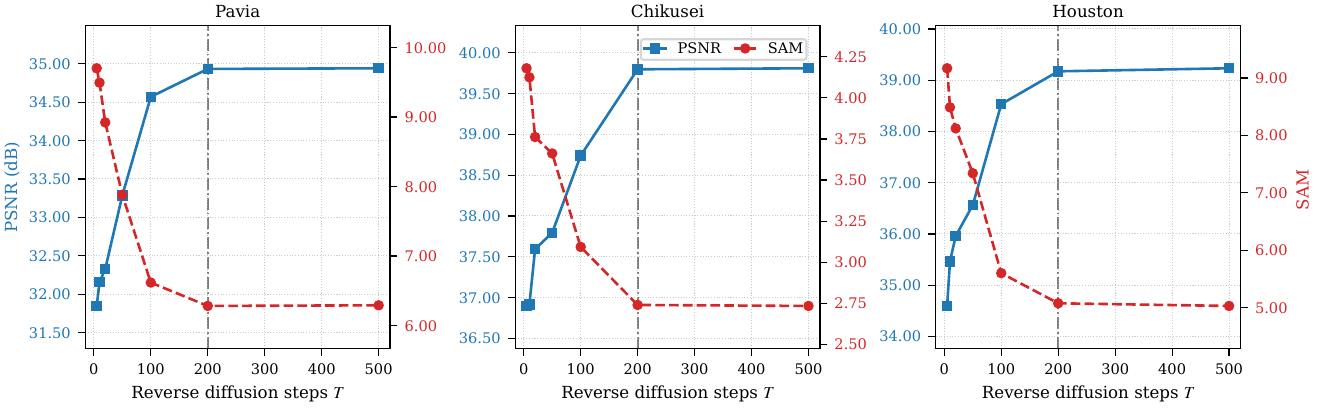}
    \caption{Effect of the number of reverse diffusion steps on Pavia, Chikusei, and Houston under the RR reference-based setting. The vertical dashed line marks the adopted setting $T=200$.}
    \label{fig:sampling_steps_all}
\end{figure*}

\noindent\textbf{Effect of Reverse Diffusion Steps.} We next vary the number of reverse diffusion steps $T$ while keeping the remaining settings fixed. Since this experiment examines the reconstruction and spectral saturation of the sampling process across datasets, Fig.~\ref{fig:sampling_steps_all} reports PSNR and SAM as representative RR metrics on Pavia, Chikusei, and Houston, with the vertical dashed line marking the adopted value $T=200$. The curves exhibit a clear transition from rapid improvement at small $T$ to saturation around $T=200$. (a) With only a few reverse steps, the diffusion prior cannot sufficiently refine spatial--spectral details; on Houston, increasing $T$ from 5 to 100 raises PSNR from 34.594 to 38.532 dB and reduces SAM from 9.171 to 5.599, with similar trends on Pavia and Chikusei. (b) Beyond $T=200$, additional refinement yields only marginal gains: increasing $T$ from 200 to 500 improves PSNR by merely 0.009, 0.013, and 0.060 dB on Pavia, Chikusei, and Houston, respectively, while the corresponding SAM changes are also limited. (c) Because sampling cost increases with the reverse trajectory length, $T=200$ retains nearly all attainable reconstruction improvement without incurring the substantially higher cost of longer sampling and is therefore adopted as a practical balance between spatial--spectral refinement and sampling efficiency. Together with the iteration analysis above, these results justify the adopted computational budgets at both the optimization and sampling levels.

\section{Conclusion}\label{sec:con}

This paper presented DIDM, a dual-modality image-prompted diffusion model for zero-shot hyperspectral pansharpening. DIDM internalized LRHS-derived spectral evidence and PAN-derived spatial structures as modality-specific prompts and injected them into a remote-sensing pre-trained diffusion prior through cross-attention, allowing the observations to participate directly in diffusion feature evolution. PAN-guided WPATV further complemented the observation-fidelity terms by imposing gradient-adaptive regularization on the spatial component, encouraging PAN-aligned structural transitions while suppressing unnecessary fluctuations in homogeneous regions.

Under the RR reference-based protocol, DIDM achieved the best values across all six metrics on Pavia, Chikusei, and Houston. Its PSNR reached 34.932, 39.797, and 39.172 dB, exceeding the strongest competing results by 1.166, 0.768, and 0.737 dB, respectively. On the released FR1 full-resolution sample, DIDM achieved the highest HQNR of 0.857, with $D_{\lambda}=0.040$ and $D_S=0.108$, and outperformed the other diffusion-prior methods on both distortion indices. The content-free prompt control indicated that the gains arose primarily from observation-aware conditioning rather than additional prompt-injection capacity alone. Moreover, the modality-specific response maps revealed distinct structure-oriented and spectral-region-aware behaviors, while the WPATV edge analysis showed improved PAN-aligned edge preservation and fewer spurious responses. Together, these results supported both the effectiveness and the design rationale of DIDM.

The current implementation still requires per-pair optimization and iterative reverse diffusion, which limits efficiency when processing large image collections despite the diminishing returns identified in both computational-budget analyses. Future work will investigate accelerated or amortized adaptation, more efficient reverse sampling, data-dependent prompt balancing, and evaluation on broader real full-resolution sensor datasets. Extending the proposed observation-aware diffusion conditioning to other hyperspectral fusion and restoration tasks also constitutes a promising direction.

% \newpage

 

% \end{IEEEbiography}



\bibliographystyle{IEEEtran}
\bibliography{DMIP_references}

\vfill

\end{document}